\pdfoutput=1

\documentclass[11pt]{article}

\usepackage[final]{acl}

\usepackage{times}
\usepackage{latexsym}

\usepackage[T1]{fontenc}

\usepackage[utf8]{inputenc}

\usepackage{microtype}

\usepackage{inconsolata}

\usepackage{graphicx}

\usepackage{relsize}
\usepackage{colortbl}
\usepackage{booktabs}
\usepackage{tabularx} 
\usepackage{lipsum}   
\usepackage{array}
\usepackage{threeparttable}
\usepackage{makecell}
\usepackage{booktabs}
\usepackage{amsmath}
\usepackage{multirow}
\usepackage{pifont}
\usepackage{enumitem}
\usepackage{hyperref} 
\title{Adversarial Attacks Against Automated Fact-Checking: A Survey}

\author{
 \textbf{Fanzhen Liu\textsuperscript{1,2}},
 \textbf{Alsharif Abuadbba\textsuperscript{2}},
 \textbf{Kristen Moore\textsuperscript{2}},
 \textbf{Surya Nepal\textsuperscript{2,3}},
\\
 \textbf{Cecile Paris\textsuperscript{2}},
 \textbf{Jia Wu\textsuperscript{1}},
 \textbf{Jian Yang\textsuperscript{1}},
 \textbf{Quan Z. Sheng\textsuperscript{1}}
\\
\\
 \textsuperscript{1}School of Computing, Macquarie University, Australia
 \\
 \textsuperscript{2}CSIRO’s Data61, Australia; 
 \textsuperscript{3}UNSW Sydney, Australia
\\
 \texttt{\{fanzhen.liu, jia.wu, jian.yang, michael.sheng\}@mq.edu.au}\\
 \texttt{\{sharif.abuadbba, kristen.moore, surya.nepal, cecile.paris\}@data61.csiro.au}
}

\begin{document}
\maketitle
\begin{abstract}

\end{abstract}
In an era where misinformation spreads freely, fact-checking (FC) plays a crucial role in verifying claims and promoting reliable information. While automated fact-checking (AFC) has advanced significantly, existing systems remain vulnerable to adversarial attacks that manipulate or generate claims, evidence, or claim-evidence pairs. These attacks can distort the truth, mislead decision-makers, and ultimately undermine the reliability of FC models. Despite growing research interest in adversarial attacks against AFC systems, a comprehensive, holistic overview of key challenges remains lacking. These challenges include understanding attack strategies, assessing the resilience of current models, and identifying ways to enhance robustness. This survey provides the first in-depth review of adversarial attacks targeting FC\footnote{Resources are available on GitHub: \href{https://github.com/FanzhenLiu/Awesome-Automated-Fact-Checking-Attacks}{https://github.com/\allowbreak FanzhenLiu/Awesome-Automated-Fact-Checking-Attacks}.}, categorizing existing attack methodologies and evaluating their impact on AFC systems. Additionally, we examine recent advancements in adversary-aware defenses and highlight open research questions that require further exploration. Our findings underscore the urgent need for resilient FC frameworks capable of withstanding adversarial manipulations in pursuit of preserving high verification accuracy.


\section{Introduction}
\label{sec:intro}
In today’s open online environment, where misinformation spreads rapidly and at scale, detecting and countering misleading claims has become a critical cybersecurity and societal challenge \cite{wu2019misinfo}. Fact-checking\footnote{We use ``fact-checking'' as a unified term encompassing both “fact verification” and related terminologies used in prior literature.} (FC) plays a central role in this effort by evaluating the veracity of claims based on accessible and trustworthy evidence.
Empirical evidence underscores the value of FC in practice. For example, even among individuals highly distrustful of fact-checkers, exposure to warning labels significantly reduces both belief in false claims and their likelihood of being shared \cite{martel2024nature}. These findings highlight the importance of robust fact-checking mechanisms in limiting misinformation spread and supporting public trust \cite{walter2020fact,guo-etal-2022-survey,augenstein2024factuality}.
\begin{figure}
    \centering
     \includegraphics[width=1\linewidth]{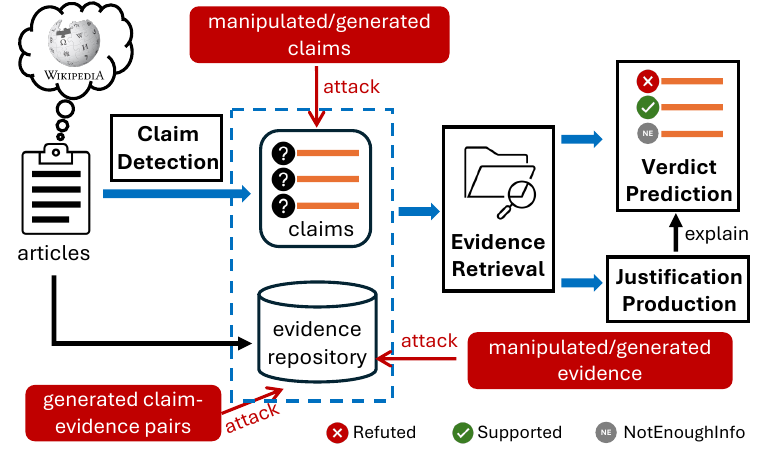}
     \caption{Overview of adversarial attacks against AFC.}
     \label{fig:FC_attack}
     \vspace{-3mm}
 \end{figure}

However, the scale, complexity, and evolving nature of online information have outpaced human capacity to fact-check content manually, as detailed in Appendix~\ref{appx:manual2auto}. This has led to the growing development of automated fact-checking (AFC) systems, which typically follow a standard four-stage pipeline, progressing from claim detection to justification production
(see Fig.~\ref{fig:FC_attack} and Sec.~\ref{sec:background}). Recent advancements in AFC cover a broad spectrum of modalities (textual and multimodal), languages (monolingual and multilingual), and data conditions \cite{guo-etal-2022-survey,akhtar-etal-2023-multimodal,gupta-srikumar-2021-x}. Despite this progress, \textit{relatively limited attention has been devoted to evaluating the adversarial robustness of these models—a critical concern as malicious actors increasingly seek to manipulate FC systems}.

In practice, adversaries may craft purpose-built attacks to obscure truth and mislead decision-making \cite{abdelnabi2023fact,atanasova-etal-2020-generating}. These attacks can undermine AFC pipelines in at least three key ways (see Fig.~\ref{fig:FC_attack}):
(1) \textit{Adversarial claim attacks}: Modifying or synthesizing misleading claims (e.g., paraphrasing or multi-hop claims) that elicit incorrect verdicts when checked against the original evidence \cite{thorne-etal-2019-evaluating,factual2025mamta};
(2) \textit{Adversarial evidence attacks}: Injecting manipulated or fabricated evidence into the corpus to mislead the retrieval process or verdict prediction \cite{du_2022_synthetic,abdelnabi2023fact}; and
(3) \textit{Adversarial claim-evidence pair attacks}: Generating synthetic pairs that maintain the original claim-evidence relationship superficially, while embedding contradictory or misleading content that confuses models trained only on truthful data \cite{schuster2019towards}.

These risks are becoming increasingly salient given recent shifts in content moderation practices. For example, Meta has discontinued its partnerships with professional human fact-checkers and transitioned to a community-driven model similar to “Community Notes.”\footnote{https://about.fb.com/news/2025/01/meta-more-speech-fewer-mistakes/} While such changes aim to democratize content moderation, they also place greater reliance on scalable, automated FC systems which must be robust to adversarial manipulation in open environments.

To build trustworthy AFC systems, resilience against such attacks is essential. Proactively identifying vulnerabilities and designing adversary-aware FC models is key to ensuring their reliability in real-world settings \cite{abdelnabi2023fact,factual2025mamta}.
Although recent years have seen increasing attention to this area, adversarial attacks against AFC systems remain underexplored. While isolated studies have investigated specific attack vectors or datasets, there is a lack of holistic understanding of how diverse adversarial strategies affect the end-to-end AFC pipeline. Several open questions remain:

\begin{enumerate}
\item What are the state-of-the-art adversarial techniques targeting the AFC pipeline? (Sec.~\ref{sec:methodology})

\item How well have current defenses developed in response to existing attacks? (Sec.~\ref{sec:defenses})

\item What are future directions for resilient, attack-aware AFC systems? (Sec.~\ref{sec:challenges})

\end{enumerate}


\noindent
\textbf{Comparison with existing surveys.}
While most existing surveys on FC focus on system development under benign conditions---such as retrieval and verification techniques~\cite{bekoulis2021review}, 
explainable FC \cite{kotonya-toni-2020-explainable}, multimodal FC \cite{akhtar-etal-2023-multimodal}, scientific FC \cite{vladika-matthes-2023-scientific}, justification generation \cite{eldifrawi-etal-2024-automated}, and LLM-assisted FC \cite{vykopal2024generative}---few consider security threats. Notably, \cite{abdelnabi2023fact} provides a detailed taxonomy of evidence manipulation attacks targeting AFC systems.  

However, to our knowledge, no existing study has comprehensively reviewed the full spectrum of adversarial attacks across the AFC pipeline---including claim manipulation, evidence poisoning, and pairwise attacks---nor proposes a unified taxonomy that spans attack targets and levels of edit granularity. This paper addresses this critical gap by systematically examining adversarial attacks directly targeting key modules (e.g., evidence retrieval and verdict prediction) in AFC systems, while excluding related but distinct tasks such as fake news detection \cite{wang2023attacking}.
Because AFC involves claim-specific evidence identification from large corpora, we exclude adjacent tasks such as textual entailment \cite{jin2020is}, natural language inference \cite{zhang2020adversarial}, and general text classification \cite{clef-checkthat:2024:task6}, which lack the full AFC pipeline.

Our main contributions are as follows:
\begin{itemize}
\item We provide the first systematic review of adversarial attacks targeting AFC systems.

\item We develop a novel attacker taxonomy that categorizes attack strategies and edit granularity, offering a framework for evaluating AFC robustness under adversarial conditions.

\item We identify key challenges in building resilient, attack-aware AFC systems, and outline promising research directions to guide future advancements in this emerging area.
\end{itemize}

\section{Background \& Preliminaries}
\label{sec:background}

Automated fact-checking (AFC) aims to verify check-worthy claims using relevant information drawn from evidence resources. Using FEVER \cite{thorne-etal-2018-fever}, one of the most widely studied AFC benchmarks, as an example, the final verdict for a claim is typically classified as Supported (SUP), Refuted (REF), or NotEnoughInfo (NEI). The core AFC pipeline comprises four major tasks \cite{guo-etal-2022-survey}:

\vspace{1mm}
\noindent
\textbf{Claim Detection.} The pipeline begins with identifying claims that are worth fact-checking. This involves assessing both the claim's verifiability (i.e., whether the claim is specific and checkable) and its potential impact (i.e., whether misinformation could cause harm or shape public opinion) \cite{guo-etal-2022-survey}.

\vspace{1mm}
\noindent
\textbf{Evidence Retrieval.} Next, the system searches associated sources to retrieve relevant evidence that supports or refutes the identified claim. The quality and alignment of this evidence strongly influence the downstream prediction of claim veracity \cite{eldifrawi-etal-2024-automated}. For example, for the claim ``The Eiffel Tower is the tallest structure in France,'' a robust system should retrieve authoritative sources clarifying that while it was once the tallest, other buildings like Tour First now surpass it. Sec.~\ref{subsec:adver_evidence} discusses how adversarial evidence can mislead retrieval and verdict stages.

\vspace{1mm}
\noindent
\textbf{Verdict Prediction.} This stage classifies the claim based on retrieved evidence. Most systems use a binary (True/False) or ternary (SUP/REF/NEI) classification scheme \cite{thorne-etal-2018-fever}, though some adopt more fine-grained verdict labels reflecting degrees of truthfulness \cite{alhindi-etal-2018-evidence,augenstein-etal-2019-multifc}. Adversarial attacks may target this step  either by introducing misleading claims (see Sec.~\ref{subsec:adver_claim}) or by distorting evidence (see Sec.~\ref{subsec:adver_evidence}) to trigger misclassification.

\vspace{1mm}
\noindent
\textbf{Justification Production.} 
Finally, some AFC systems provide textual justification for their verdicts. These explanations clarify how evidence supports or refutes the claim and may incorporate reasoning, common sense inference, or attribution of assumptions \cite{guo-etal-2022-survey,eldifrawi-etal-2024-automated}. Justifications enhance not only transparency but also trustworthiness and persuasive power—particularly important in adversarial contexts \cite{zeng-etal-tacl-2024,he2025fact2fiction}.

Figure~\ref{fig:FC_attack} illustrates the core AFC pipeline and highlights how adversarial manipulations of claims, evidence, or claim-evidence pairs can compromise each stage.



\section{Overview of Adversarial Attacks}
\label{sec:FC_attack}
This section provides a structured overview of off-the-shelf adversarial attacks targeting AFC systems. We categorize attacks into three main types based on which component of the AFC pipeline they target: \textit{Adversarial claim attacks}, \textit{Adversarial evidence attacks}, and \textit{Adversarial claim-evidence pair attacks} (discussed in Sec.~\ref{sec:intro}). 
Each category is further broken down based on the information source available to the attacker (e.g., original claims, evidence repositories, or open corpora). Full details of these methods and settings are provided in Tables~\ref{tab:adv_claim}–\ref{tab:adv_evidence_claim_pair} in Appendix~\ref{appx:tables}.

\noindent
\textbf{Adversarial Claim Attacks.}
As shown in Fig.~\ref{fig:adv_claim} in Appendix~\ref{appx:figures}, adversarial claim attacks aim to fool the verification model by supplying manipulated or newly generated claims, mostly assuming black-box access to the verdict prediction module. Based on the source of information used, such attacks fall into two types: (1) \textit{Evidence-guided:} Generating claims by editing or recomposing sentences from existing evidence documents; and (2) \textit{Claim-guided:} Manipulating original claims (e.g., via paraphrasing or adversarial transformation) to induce misclassification.
Adversarial claims are typically crafted 
using either rule-based transformations or generative language models (LMs). 



\noindent
\textbf{Adversarial Evidence Attacks.}
Figure~\ref{fig:adv_evidence} in Appendix~\ref{appx:figures} presents the structure of adversarial evidence attacks, which aim to inject misleading or distracting evidence into the retrieval corpus, causing downstream retrieval or prediction errors. With black-box access to the verification model, attackers can exploit four types of guidance: (1) \textit{Evidence-guided}: Editing or modifying gold evidence; (2) \textit{Claim-guided}: Generating misleading evidence directly from the target claim; (3) \textit{Open corpus-guided}: Retrieving unrelated but plausible-looking evidence from an external corpus; (4) \textit{Retrieval-guided}: Manipulating the evidence after it is retrieved for a target claim; and (5) \textit{Justification-guided:} Crafting malicious evidence by leveraging the justification associated with a target claim.
The attacks use either rule-based editing or LM-based generation. 

\noindent
\textbf{Adversarial Claim-Evidence Pair Attacks.}
Unlike the aforementioned categories that target individual pipeline components, claim-evidence pair attacks generate synthetic pairs that resemble valid claim-evidence relationships from the original dataset but encode contradictory or misleading content. 
Pairs are generated using either rule-based techniques or large language models (LLMs), as illustrated in Figure~\ref{fig:adv_pair} in Appendix~\ref{appx:figures}. 


\section{Taxonomy of Adversarial Attacks}
\label{sec:methodology}
To make sense of the diverse and rapidly growing space of adversarial attacks against AFC systems, we propose a unified, technique-centric taxonomy. While previous surveys have primarily focused on the design of FC systems including dataset construction, model architectures, and pipeline components, relatively \textit{little attention has been paid to adversarial methods that actively challenge these systems.} Our goal is to synthesize existing attack techniques through the lens of their operational behavior, offering a structured understanding of how and where they compromise the AFC pipeline.

\begin{figure*}[htbp]
    \centering
    \includegraphics[width=.95\linewidth]{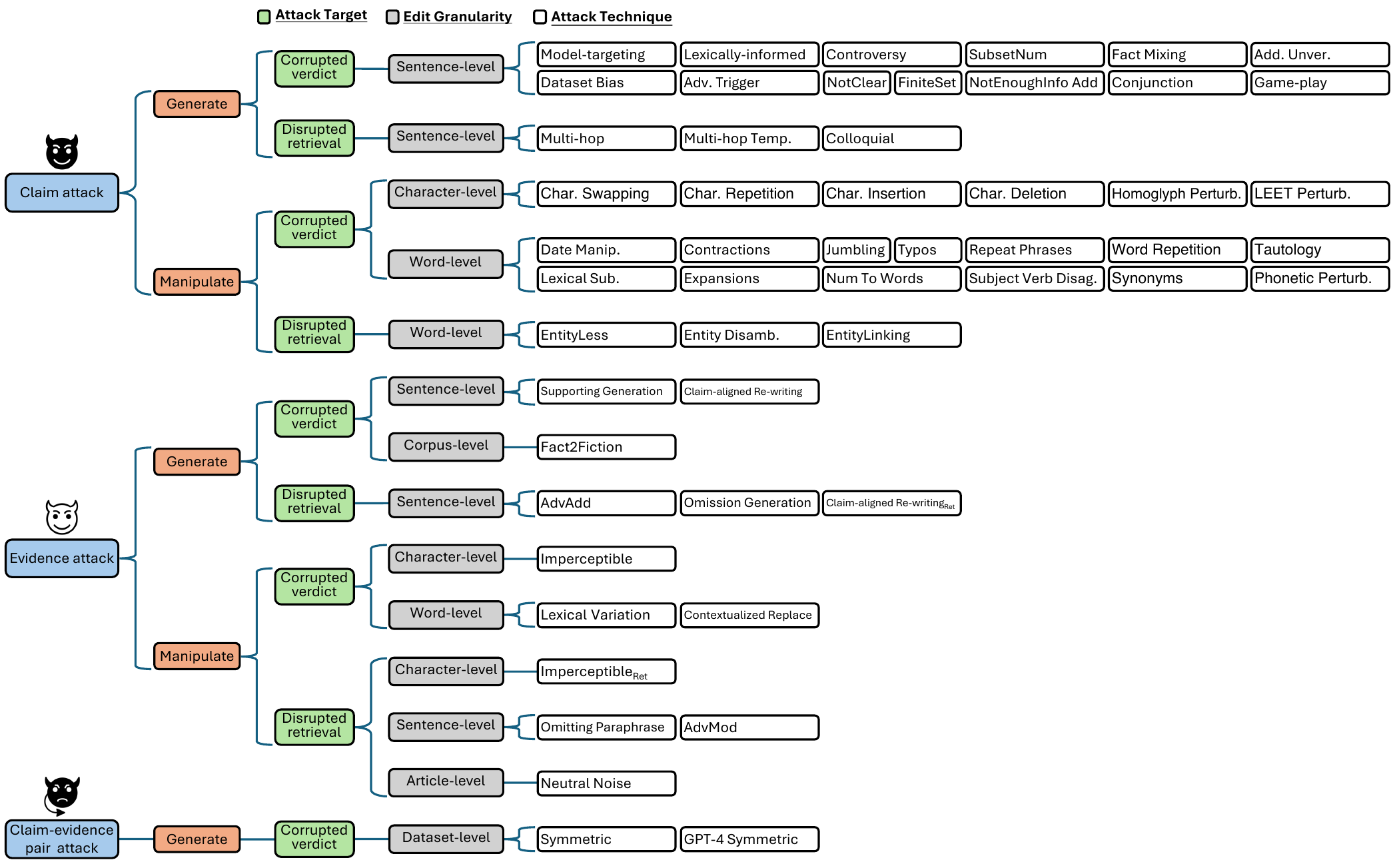}
    \caption{A technical taxonomy of adversarial attacks against AFC.}
    \label{fig:taxonomy}
    \vspace{-3mm}
\end{figure*}

Specifically, we organize attacks based on two dimensions: (1) \textit{attack target} — identifying the specific component of the AFC pipeline being compromised (i.e., verdict prediction or evidence retrieval), which aligns closely with system-level vulnerabilities; and (2) \textit{edit granularity} — specifying the structural level at which adversarial perturbations are applied,  ranging from subtle character-level edits to full sentence or article modifications, which influences both detectability and generalization performance. This taxonomy allows us to systematically compare attack strategies across diverse components of the AFC pipeline and to highlight trends in their implementation, effectiveness, and robustness implications.
Fig.~\ref{fig:taxonomy} provides an overview of this taxonomy, which we elaborate on in the following subsections. Corresponding empirical results are discussed in Sections~\ref{subsec:adver_claim}--\ref{subsec:pairs}.

Besides the taxonomy, we provide structured summaries of each attack type in Appendix~\ref{appx:tables}, reporting key characteristics including attack target, method, and target dataset, allowing 
a deeper comparative analysis across techniques.
Other factors such as model access (black-box vs. white-box), automation strategy (rule-based or LM-based), and semantic preservation, are reported less consistently across studies and are less suitable as primary organizational axes. 
We include these dimensions in our 
comparison (Tables~\ref{tab:adv_claim}--\ref{tab:adv_evidence_claim_pair} in Appendix~\ref{appx:tables}), which provides a more fine-grained analysis across attacks. We consider these dimensions complementary to our taxonomy, and promising candidates for future extensions or alternative schemes.




Next, we delve into the three categories outlined in the taxonomy—Adversarial Claim Attacks, Adversarial Evidence Attacks, and Adversarial Claim-Evidence Pair Attacks—and examine their mechanisms across attack target and edit granularity.

\section{Methodology of Adversarial Attacks}
\subsection{Adversarial Claim Attacks}
\label{subsec:adver_claim}

Adversarial Claim Attacks target the claim component of fact-checking pipelines by either generating new claims or manipulating existing ones to mislead FC models. Among the 38 surveyed attacks, 16 focus on sentence-level generation, while 22 involve manipulations at the character or wordlevel, as shown in Fig.~\ref{fig:taxonomy}. Most studies evaluate attacks using the FEVER 1.0/2.0 benchmark \cite{thorne-etal-2018-fever, thorne-etal-2019-fever2}, but a universal evaluation framework—with standardized models, metrics, and perturbation budgets—is still lacking.

\subsubsection{Generation-Based Attacks} 
\textit{Rule-Based and Game-Inspired Generation.} Rule-based strategies, including Model-targeting (with Semantically Equivalent Adversarial Rules (SEARs)) and Dataset Bias, generate adversarial claims that preserve semantic meaning while inducing misclassifications in FC models \cite{thorne-etal-2019-evaluating, ribeiro-etal-2018-semantically}. These approaches reveal several key insights. First, the Dataset Bias attack outperforms other rule-based methods by producing grammatically coherent, label-consistent claims that significantly degrade model performance—particularly in systems like Papelo \cite{malon-2018-team} and Enhanced ESIM \cite{hanselowski-etal-2018-ukp}. Second, attacks such as Controversy, NotClear, SubsetNum, and Multi-hop introduce ambiguity, underspecification, or the need for multi-hop reasoning, thereby challenging a model's ability to generalize beyond surface-level heuristics \cite{kim-allan-2019-fever, hidey-etal-2020-deseption}. Third, the Game-play attack enhances diversity and realism by involving human annotators in competitive settings to craft difficult yet plausible claims, further broadening the adversarial landscape \cite{eisenschlos-etal-2021-fool}.

\textit{Language Model (LM)-Assisted Generation and Colloquial Claims.}
Language models such as GPT-2 \cite{radford2019language} have been leveraged to generate adversarial claims through various techniques. In the Lexically-informed attack, claims are paraphrased using lexical modifications to subtly alter input semantics while preserving fluency \cite{thorne-etal-2019-evaluating}. The Adv. Trigger attack embeds specific triggers into input text to flip model predictions with minimal edits, maintaining both grammaticality and semantic coherence—particularly effective when shifting verdicts from ``Supported'' to ``Refuted'' \cite{atanasova-etal-2020-generating}. The Fact Mixing attack employs GPT-2 to generate plausible but deceptive claims by blending facts from multiple sources, thereby misleading AFC systems without compromising fluency or label consistency \cite{niewinski-etal-2019-gem}. In a related vein, the Colloquial attack rephrases claims into informal, conversational language using a BART model fine-tuned on open-domain dialogue datasets \cite{lewis-etal-2020-bart}. This stylistic transformation significantly degrades document retrieval performance; for example, the WikiAPI retriever's recall drops from 90\% on original FEVER claims to 72.2\% on their colloquial counterparts \cite{kim-etal-2021-robust}.

\subsubsection{Manipulation-Based Attacks} 
Beyond generation, many attacks manipulate existing claims at different levels of granularity. At the \textit{character level}, subtle perturbations such as character swapping, deletions, insertions, homoglyph substitutions (from the Unicode security dictionary), and LEET-style transformations can corrupt token representations and mislead AFC models \cite{factual2025mamta}.
At the \textit{word level}, manipulations alter semantics while maintaining surface plausibility. Examples include date changes, subject-verb agreement errors \cite{sai-etal-2021-perturbation}, synonym and hypernym substitutions, and linguistic variations like contractions, expansions, word order jumbling, number-to-word conversions, and phrase repetition. Under the FactEval benchmark \cite{factual2025mamta}, attacks like Typos, Tautology, and Phonetic Perturb. prove especially disruptive to both traditional and LLM-based fact-checkers.
\textit{Entity-focused} word-level attacks target retrieval. The EntityLess attack replaces specific entity names with generic terms (e.g., ``Harvard University'' becomes ``university''), EntityLinking substitutes names with uncommon aliases, and Entity Disamb. introduces ambiguity—all of which degrade retrieval precision and lead to incorrect or incomplete verdicts \cite{kim-allan-2019-fever}.

\textit{Evaluation and Impact.} Several manipulation-based attacks demonstrate high effectiveness while preserving claim plausibility. SubsetNum results in an average of over 80\% incorrect verdict predictions across tested FC systems \cite{hidey-etal-2020-deseption}, while both SubsetNum and  Fact Mixing cause systems to fail in evidence retrieval entirely, yielding zero FEVER scores despite correct labeling \cite{thorne-etal-2019-fever2}. Similarly, Adv. Trigger maintains grammaticality yet causes significant performance degradation \cite{alzantot-etal-2018-generating}.

\textit{LLM Robustness under FactEval.} The FactEval benchmark evaluates LLMs against 17 structured manipulations. Among the tested models, Gemma 7B Instruct \cite{gemmateam2024} demonstrates the best generalization in zero- and few-shot settings, yet all models including BERT \cite{devlin-etal-2019-bert}, Liama3 \cite{grattafiori2024llama3}, and Mistral \cite{jiang2023mistral7b} remain vulnerable to minor edits. Liama3 exhibits the highest attack success rates, while Mistral performs better under chain-of-thought prompting. These results underscore the fragility of even instruction-tuned models when exposed to benign-looking perturbations.

\vspace{1mm}
\noindent
\textbf{Remark 1.} Most adversarial claim attacks operate in black-box settings. Model-targeting leveraging FC model predictions \cite{Nie_Chen_Bansal_2019}, suggests a need for exploring white-box scenarios. Notably, improved retrieval does not guarantee better verdict predictions, especially in complex, multi-hop or blended claims. Attacks such as Multi-hop Temp., Colloquial, and Fact Mixing reveal that many models rely on shallow heuristics rather than deep reasoning. Although \citet{factual2025mamta} have recently introduced FactEval, a structured benchmark for LLM evaluation, further work is needed to test robustness against more diverse, rule-based and generation-driven adversarial strategies. For an attack to be considered effective, it must not only reduce model performance but also preserve fluency and label consistency, so that failures reflect reasoning flaws rather than noise sensitivity.

\subsection{Adversarial Evidence Attacks}
\label{subsec:adver_evidence}
Adversarial evidence attacks, in contrast to adversarial claim attacks, generate new or modify existing evidence articles without changing claims. 
The 13 identified attacks in Fig.~\ref{fig:taxonomy} aim to mislead the verifier into incorrect verdicts or disrupt evidence retrieval for original claims. 

\subsubsection{Generation-Based Attacks} 
Most adversarial evidence attacks employ \textit{sentence-level} generation. 
To corrupt verdicts on target claims, the Claim-aligned Re-writing attack leverages a T5 model \cite{raffel2020exploring} to reconstruct context-preserving supporting evidence that flips REF verdicts, but it is hard to apply to NEI claims. In contrast, the Supporting Generation attack fine-tunes GPT-2 to create supporting evidence that boosts the SUP probability of BERT-based stance prediction, thereby misleading verification on both NEI and REF claims \cite{abdelnabi2023fact}. 

Other generation-based attacks compromise evidence retrieval by removing or altering content from gold evidence. 
The AdvAdd attack, which fabricates evidence using the Grover disinformation generator \cite{zellers2019defending}, shows that verification accuracy drops and prediction shifts increase as more poisoned evidence is added \cite{du_2022_synthetic}. NEI claims are especially vulnerable, despite similar contamination rates with REF claims. 
The Omission Generation attack removes sentences or optional constructs (e.g., prepositional phrases and modifiers) from gold evidence associated with SUP/REF claims, impairing evidence sufficiency prediction in FC models—BERT \cite{devlin-etal-2019-bert}, RoBERTa \cite{liu2019roberta}, and ALBERT \cite{lan2020ALBERT}—particularly for adverbial omissions hardest to detect, while date omissions are easiest \cite{atanasova2022fact}.
Empirically, the Supporting Generation outperforms the AdvAdd-based attack using only 10\% of training data, as smaller subsets produce more direct, claim-supporting sentences, while larger datasets create less misleading, diverse outputs \cite{abdelnabi2023fact}.

Fact2Fiction marks the initial effort in FC attacks targeting agentic FC systems, adopting \textit{corpus-level} generation to mimic claim decomposition and to exploit system-generated justifications in creating malicious evidences that compromise sub-claim verification \cite{he2025fact2fiction}. Its success against two state-of-the-art agentic FC systems \cite{braun2025defame,rothermel-etal-2024-infact} underscores the urgent need for defensive strategies.

\subsubsection{Manipulation-Based Attacks} 
\textit{Character-level} manipulation relies on disrupted tokenization by altering evidence words with homoglyphs, invisible, or deletion control characters \cite{boucher2022bad}. This Imperceptible attack evolutionarily optimizes manipulations to reduce the BERT classifier's probability of the correct claim label \cite{devlin-etal-2019-bert}, serving to corrupt verdict prediction. 
Instead of directly misleading final verdicts, the Imperceptible\textsubscript{Ret} attack reduces evidence retrievability by targeting entity mentions.

\textit{Word-level} manipulations are preformed assisted with LMs.
The Lexical Variation attack \cite{alzantot-etal-2018-generating} produces adversarial claim-paired evidence that successfully fools a pre-trained model like RoBERTa\textsubscript{BASE} \cite{liu2019roberta} and is generally less likely to be retrieved \cite{abdelnabi2023fact}. Seeking more fluent and high-quality perturbations, the Contextualized Replace attack built on \cite{li-etal-2020-bert-attack} uses a pre-trained BERT masked model to obtain candidate replacements for salient words in evidence.
Despite a low retrieval rate for manipulated evidence, this attack is effective in reversing the verdict of SUP/REF claims.


Some \textit{sentence-level} manipulations address issues where attacks hide evidence from FC models but not from human readers, or introduce syntactic errors. For example, Omitting Paraphrase uses PEGASUS~\cite{pmlr-v119-zhang20ae} to paraphrase evidence while omitting claim-salient snippets, reducing retrievability~\cite{abdelnabi2023fact}—KGAT (BERT\textsubscript{BASE}) mistakenly retrieves 54.4\% of adversarial envidence, flipping SUP/REF to NEI.
AdvMod also leverages PEGASUS, appending a paraphrased claim to evidence to confuse retrievers \cite{du_2022_synthetic}. It disrupts REF classification in CorefBERT~\cite{ye-etal-2020-coreferential} and MLA~\cite{kruengkrai-etal-2021-multi} more than NEI.
At the \textit{article level}, Neutral Noise adds top Bing search results with high BM25 scores and neutral entailment~\cite{samarinas2021improving}, significantly degrading both sparse (BM25) and dense retriever performance~\cite{samarinas2020latent}.

\vspace{1mm}
\noindent
\textbf{Remark 2.} 
The attacks in \cite{abdelnabi2023fact} (e.g., Omitting Paraphrase and Imperceptible) were tested with both white-box and black-box access to the FC retriever, showing a thorough examination of both settings for other adversarial evidence attacks.
Empirical study \cite{abdelnabi2023fact} has revealed vulnerabilities in FC model, particular the susceptibility to the absence of alternative or competing evidence (AdvAdd),
despite some resilience to minimal refuting evidence. This emphasizes the importance of comprehensive evidence retrieval in defending against adversaries. 
%
LLMs facilitate more fluent and sophisticated evidence manipulation, making attacks like Contextualized Replace highly disruptive to verdict prediction, but other LLM-assisted attacks like Supporting Generation may introduce errors and misinterpretations in complex cases like negation.
%
%
Attacks like Imperceptible\textsubscript{Ret} and Claim-aligned Re-writing\textsubscript{Ret} select the adversarial evidence based on claim-evidence agreement, but generally achieve less degradation in overall verification compared to evidence selected for claim misclassification.  
This raises the question of how strongly verdict prediction truly depends on evidence quality.


\subsection{Adversarial Claim-Evidence Pair Attacks}
\label{subsec:pairs}
Distinct from the two previously described classes of attacks, adversarial attacks can generate claim-evidence pairs by leveraging natural biases present in current benchmark FC datasets, thereby making it hard for pre-trained FC models to produce correct verdicts on these new pairs.

\subsubsection{Generation-Based Attacks}
\citet{schuster2019towards} exposed spurious correlations between claim patterns and veracity, i.e., idiosyncratic biases, arising from the manual construction of datasets like FEVER 1.0 \cite{thorne-etal-2018-fever}. Building on this, the Symmetric attack manually constructs synthetic claim-evidence pairs that retain the original relational label (SUPPORTS or REFUTES) while introducing contradictory factual content. Combined with FEVER 1.0, this yields two additional cross pairs with inverse labels, forming the unbiased FEVER-sym dataset for more rigorous evaluation. Extending to Chinese, \citet{zhang2024need} examined domain and cultural biases in CHEF \cite{hu-etal-2022-chef}, revealing limitations of translation-based and multilingual LMs. Replacing manual construction with an LLM-driven approach, the GPT-4 Symmetric attack uses GPT-4 \cite{openai2024gpt4technicalreport} to automatically generate analogous claim-evidence pairs.

\textit{Evaluation and Impact.}
The adversarial pairs manually constructed by the Symmetric attack result in a significant degradation in FC verification accuracy, with performance falling below 60\%—a decline of over 20 points \cite{schuster2019towards}. GPT-4 further demonstrates its ability to generate more challenging pairs. For example, the accuracy of Chinese DeBERTa \cite{zhang2023fengshenbang}, a strong Chinese-specific model, decreases from 86.69\% on CHEF to 57.84\% on GPT-4-generated pairs \cite{zhang2024need}. These sharp performance gaps underscore the models’ reliance on surface-level cues and expose biases inherent in dataset construction.
%


\vspace{1mm}
\noindent
\textbf{Remark 3.}
Attacks that generate claim-evidence pairs, either manually or using GPT-4 to construct unbiased datasets, are conducted with no access to both the FC verification and retrieval models.
Since FC models are typically pre-trained on popular benchmark datasets like FEVER 1.0, they often inherit the idiosyncratic biases embedded in these datasets. Introducing new unbiased datasets can help develop FC models that focus more on underlying semantic relationships rather than superficial cues like specific entities or objects.

\section{Defending Against Adversarial Attacks}
\label{sec:defenses}
Table~\ref{tab:adv_defense} summarizes FC defense strategies, outlining the adversarial attacks and datasets they target, to assess the coverage of current defenses.

\begin{table}[ht!]
\centering
\begin{threeparttable}
\resizebox{0.49\textwidth}{!}{%
\begin{tabular}{ccc}
\toprule
\textbf{Attack Technique} & \textbf{Target Dataset} & \textbf{FC Defense} \\ \midrule
\multicolumn{3}{c}{\textbf{Adversarial Claim Attacks}} \\ 
\midrule
\rowcolor[HTML]{C0C0C0} 
\begin{tabular}[c]{@{}c@{}}Model-targeting\\ Dataset Bias\\ Lexically-informed\end{tabular} & FEVER-adv & CLEVER \cite{xu2023counterfactual} \\
Conjunction & FEVER 2.0 & \citet{hidey-etal-2020-deseption} \\
\rowcolor[HTML]{C0C0C0}
Multi-hop & FEVER 2.0 & \citet{hidey-etal-2020-deseption} \\
Multi-hop & DeSePtion & AdMIRaL \cite{aly-vlachos-2022-natural} \\
\rowcolor[HTML]{C0C0C0}
Add. Unver. & FEVER 2.0 & \citet{hidey-etal-2020-deseption} \\
Date Manip. & FEVER 2.0 & \citet{hidey-etal-2020-deseption} \\
Multi-hop Temp. & FEVER 2.0 & \citet{hidey-etal-2020-deseption} \\
\rowcolor[HTML]{C0C0C0}
Entity Disamb. & FEVER 2.0 & \citet{hidey-etal-2020-deseption} \\
\rowcolor[HTML]{C0C0C0}
Lexical Sub. & FEVER 2.0 & \citet{hidey-etal-2020-deseption} \\
\midrule
\multicolumn{3}{c}{\textbf{Adversarial Evidence Attacks}} \\ 
\midrule
\rowcolor[HTML]{C0C0C0}
Neutral Noise & Factual-NLI+ & Quin+ \cite{samarinas2021improving} \\
\multirow{2}{*}{Omission Generation} & \multirow{2}{*}{SufficientFacts} & CL \cite{atanasova2022fact} \\
&  &  CAD \cite{atanasova2022fact} \\
\midrule
\multicolumn{3}{c}{\textbf{Adversarial Claim-Evidence Pair Attacks}} \\ 
\midrule
\multirow{9}{*}{Symmetric} & PolitiHop-sym &Causal Walk \cite{zhang2024causal}  \\ 
\cmidrule{2-3}
& \multirow{8}{*}{FEVER-sym} & CLEVER \cite{xu2023counterfactual} \\
&  & CICR \cite{tian2022debiasing} \\
&  & CorssAug \cite{lww2021crossaug} \\
&  & \cite{ghaddar-etal-2021-end} \\
&  & DFL \cite{karimi-mahabadi-etal-2020-end} \\
&  & PoE \cite{karimi-mahabadi-etal-2020-end} \\
&  & Reweighting \cite{schuster2019towards} \\
\bottomrule
\end{tabular}%
}
\begin{tablenotes}
\tiny
\item \textbf{CL}: contrastive learning; \textbf{CAD}: counterfactually augmented data. 
\end{tablenotes}
\end{threeparttable}
\caption{Overview of FC defenses against adversarial claim, evidence, and claim-evidence pair attacks.}
\label{tab:adv_defense}
\vspace{-3mm}
\end{table}

\subsection{Defenses Against Adversarial Claims}
Current efforts have produced only three defense strategies, targeting 10 of the 38 adversarial claim attacks discussed in Sec.~\ref{subsec:adver_claim}.
CLEVER \cite{xu2023counterfactual}, a counterfactual-based method, mitigates performance drop of FC models trained on the biased FEVER 1.0 dataset \cite{thorne-etal-2018-fever} when evaluated on unbiased data. Although it does not directly tackle Model-targeting, Dataset Bias, and Lexically-informed attacks \cite{thorne-etal-2019-evaluating}, it improves verification accuracy on adversarial SUP and REF claims by around 10 points for BERT \cite{devlin-etal-2019-bert}. 
%

\citet{hidey-etal-2020-deseption} proposed a defense system tackling adversarial claims with multiple propositions posed by Conjunction, Multi-hop, and Add. Unver. attacks, employing a pointer network to rerank candidate documents and jointly predict evidence and veracity. To counter Date Manip. and Multi-hop Temp. attacks, a post-processing module for temporal reasoning is incorporated. A fine-tuned BERT further enhances its ability to handle entity ambiguity and complex lexical relations.

AdMIRaL enhances multi-hop retrieval with a retrieve-and-rerank solution that jointly scores documents and sentences with an autoregressive retriever  \cite{aly-vlachos-2022-natural}. Its natural logic-based proof system dynamically stops retrieval upon finding sufficient evidence, significantly boosting evidence retrieval on the DeSePtion dataset \cite{hidey-etal-2020-deseption}.
Other studies like \cite{xu2023counterfactual} and \cite{zhang2024causal} explore internal multi-hop reasoning for original claims, but it remains unclear if their methods effectively handle adversarial claims from the Multi-hop attack.

\subsection{Defenses Against Adversarial Evidence}
Among the 13 adversarial evidence attacks in Fig.~\ref{fig:taxonomy}, only two---Neutral Noise and Omission Generation---have been addressed by existing defence strategies. \citet{samarinas2021improving} proposed Quin+, a hybrid retriever that combines dense retriever embeddings~\cite{samarinas2020latent} with BM25-based sparse retrieval to counter the Neutral Noise attack. 
This improves ranking and early identification of relevant articles, leading to better verification accuracy on the Factual-NLI+ dataset when paired with embedding-based or sequence-labeling models.
To address the Omission Generation attack, \citet{atanasova2022fact} applied a contrastive learning objective and counterfactual data augmentation. Tested on models including BERT, RoBERTa, and ALBERT, this approach improves evidence sufficiency prediction by up to 16.83 macro-F1 points on SufficientFacts instances.\\

\subsection{Defenses Against Adversarial Claim-Evidence Pairs}
Several debiasing training paradigms have been proposed to address the Symmetric attack by reducing the impact of idiosyncratic biases in training data. 
All aim to improve verdicts for unbiased test-time claims.


\citet{schuster2019towards} reweighted the instances to flatten the correlation between the claim n-grams (e.g., \textit{did not}) and the claim labels. This method helps the best-performing tested model, BERT, achieve a 3.3-point accuracy gain on FEVER-sym claims.
%
Similarly, Debiased Focal Loss (DFL) downweights the most biased examples during training \cite{karimi-mahabadi-etal-2020-end}, while Product of Experts (PoE), inspired by \cite{hinton2002}, reduces gradient updates for claims confidently predicted by a bias model. The two strategies encourage the verification model to focus less on spurious correlations in training data, improving the verdict accuracy of BERT by over 7.5 points.

CrossAug \cite{lww2021crossaug}, a contrastive data augmentation method, combines neural-based negative claim generation with lexical search-based evidence modification, outperforming both the reweighting approach \cite{schuster2019towards} and PoE \cite{karimi-mahabadi-etal-2020-end}.
Approaching debiasing from a counterfactual view, CLEVER \cite{xu2023counterfactual}, discussed earlier, subtracts the output of a claim-only model from that of an independent claim-evidence fusion model. It yields an accuracy increase of at least 5.85 points over CrossAug and PoE variants on unbiased claims.

Causal intervention also helps mitigate idiosyncratic biases in training data—whether through CICR's counterfactual reasoning \citep{tian2022debiasing}, which outperforms PoE by 3.76 accuracy points on FEVER-sym claims, or via front-door adjustment in Causal Walk \citep{zhang2024causal} which outperforms CICR and CLEAR by over 4.97 accuracy points on the claim-evidence graph.




\section{Challenges \& Opportunities}
\label{sec:challenges}
Despite progress in adversarial attacks against AFC systems, several unresolved challenges remain, pointing to important directions for future research.

\vspace{1mm}
\noindent
\textbf{Universal evaluation benchmark.} A universal benchmark for evaluating AFC model robustness across datasets and metrics is still lacking (see Appendix~\ref{appx:evaluation}), limiting comparability across attacks and defenses.
While most adversarial attacks investigated in this work are designed to deceive AFC systems, their ability to evade human detection remains underexplored. Future work should evaluate adversarial success from both system-level and human-centric perspectives.



\vspace{1mm}
\noindent
\textbf{Stronger defenses.} 
As discussed in Sec.~\ref{sec:defenses}, current defenses address \textit{only 13 of the 53 attacks across all categories}, covering less than a quarter. More disruptive attacks exploiting inductive reasoning and knowledge compositional weaknesses (e.g., SubsetNum and Multi-hop Temp.) remain unsolved and demand stronger mitigation strategies.

\vspace{1mm}
\noindent
\textbf{Multimodal attacks.} 
Most attacks target text, overlooking real-world FC tasks that span text, images, and videos \cite{cekinel-etal-2025-multimodal,zhang2024escnet}. Future work should explore cross-modal adversaries to assess AFC robustness under more complex, multimodal threats.

\vspace{1mm}
\noindent
\textbf{Real-time attacks.} Facts change over time, making AFC tasks inherently dynamic. Attackers can exploit outdated model knowledge by crafting claims that are subtly misleading due to temporal shifts. While many attacks target linguistic patterns (e.g., Lexically-informed and EntityLess) or logical reasoning (e.g., SubNum and Multi-hop Temp.), few account for this evolving nature of truth. Addressing temporal vulnerability is critical for building robust AFC systems for practical use.

\vspace{1mm}
\noindent
\textbf{White-box verification attacks.} Attacks exploiting access to FC models for verdict prediction (i.e., white-box verification) are underexplored. Despite the practical realism of black-box access (where only prediction APIs are exposed), white-box investigation is vital given the growing prevalence of powerful open-source LM-based tools.

\vspace{1mm}
\noindent
\textbf{Vulnerability testing of LLM-based systems.} Although recent work \cite{factual2025mamta} has explored the robustness of representative LLMs for FC against handcrafted perturbations on claims, future research is expected to encompass not only adversarial claims generated by other attack techniques but also adversarial evidence and claim-evidence pairs from the other two categories.
Additionally, being aware of potential misuse of LLMs, it is critial for future work to evaluate how LLM-generated content (e.g., through specially designed prompts) affects the robustness of AFC systems. For instance, \cite{sakib-etal-2025-battling} explores how subtle prompt manipulations can induce factual inconsistencies in LLM outputs, while \cite{zou2025poisonedrag} investigates that only a few malicious texts injected into the knowledge database of a Retrieval-Augmented Generation (RAG) system could result in significant performance degradation across general NLP tasks. These attacks expose vulnerabilities in LLM reasoning, factual grounding, and knowledge governance, highlighting the need for robust fact-checking tailored to generative models.


\section{Conclusion}
\label{sec:conclusion}
This work positions a technical view of adversarial attacks against AFC systems, outlining the pipeline for crafting target-built adversarial instances (claim, evidence, or claim-evidence pairs) to interact with these systems. We introduce a new taxonomy of attack techniques, discuss their destructive effects, and examine existing defense strategies. Finally, key challenges and opportunities are highlighted to guide future research in this emerging area.

\section*{Acknowledgments}
This work was supported by the Australian Research Council Project DP230100899, Macquarie University Data Horizons Research Centre and Applied AI Research Centre.

\newpage
\section*{Limitations}
While we focus on works that directly target system-level vulnerabilities in automated fact-checking (AFC) through adversarial attacks, we exclude broader research on input manipulation or text generation for other tasks such as textual entailment, natural language inference, and general
text classification. 
Given the importance of building resilient, attack-aware AFC systems to counter real-world misinformation, our goal is to address this critical gap by providing a systematic investigation of adversarial attacks specifically targeting AFC systems. 

\section*{Ethics Statement}
This research undertakes an investigation into adversarial attacks against automated fact-checking (AFC) systems with the aim of systematically assessing their vulnerabilities and informing the development of more robust verification tools. Recognizing that our findings could inadvertently reveal methods for spreading false information, our central ethical commitment is to defense: to support the creation of attack-aware and resilient AFC systems.
The broader ethical significance of this work lies in its potential to strengthen the integrity of information ecosystems, which is essential for combating misinformation and fostering informed public discourse. We call upon future researchers to build upon our work with rigorous attention to ethical deployment, fairness across demographic groups, and transparency in system design.

\bibliography{reference}

\begin{thebibliography}{99}
\providecommand{\natexlab}[1]{#1}

\bibitem[{Abdelnabi and Fritz(2023)}]{abdelnabi2023fact}
Sahar Abdelnabi and Mario Fritz. 2023.
\newblock \href {https://www.usenix.org/conference/usenixsecurity23/presentation/abdelnabi} {{Fact-Saboteurs}: A taxonomy of evidence manipulation attacks against {Fact-Verification} systems}.
\newblock In \emph{32nd USENIX Security Symposium (USENIX Security 23)}, pages 6719--6736.

\bibitem[{Akhtar et~al.(2023)Akhtar, Schlichtkrull, Guo, Cocarascu, Simperl, and Vlachos}]{akhtar-etal-2023-multimodal}
Mubashara Akhtar, Michael Schlichtkrull, Zhijiang Guo, Oana Cocarascu, Elena Simperl, and Andreas Vlachos. 2023.
\newblock \href {https://aclanthology.org/2023.findings-emnlp.361} {Multimodal automated fact-checking: A survey}.
\newblock In \emph{Findings of the Association for Computational Linguistics: EMNLP 2023}, pages 5430--5448.

\bibitem[{Alhindi et~al.(2018)Alhindi, Petridis, and Muresan}]{alhindi-etal-2018-evidence}
Tariq Alhindi, Savvas Petridis, and Smaranda Muresan. 2018.
\newblock \href {https://doi.org/10.18653/v1/W18-5513} {Where is your evidence: Improving fact-checking by justification modeling}.
\newblock In \emph{Proceedings of the First Workshop on Fact Extraction and {VER}ification ({FEVER})}, pages 85--90.

\bibitem[{Aly and Vlachos(2022)}]{aly-vlachos-2022-natural}
Rami Aly and Andreas Vlachos. 2022.
\newblock \href {https://doi.org/10.18653/v1/2022.emnlp-main.411} {Natural logic-guided autoregressive multi-hop document retrieval for fact verification}.
\newblock In \emph{Proceedings of the 2022 Conference on Empirical Methods in Natural Language Processing}, pages 6123--6135.

\bibitem[{Alzantot et~al.(2018)Alzantot, Sharma, Elgohary, Ho, Srivastava, and Chang}]{alzantot-etal-2018-generating}
Moustafa Alzantot, Yash Sharma, Ahmed Elgohary, Bo-Jhang Ho, Mani Srivastava, and Kai-Wei Chang. 2018.
\newblock \href {https://doi.org/10.18653/v1/D18-1316} {Generating natural language adversarial examples}.
\newblock In \emph{Proceedings of the 2018 Conference on Empirical Methods in Natural Language Processing}, pages 2890--2896.

\bibitem[{Atanasova et~al.(2022)Atanasova, Simonsen, Lioma, and Augenstein}]{atanasova2022fact}
Pepa Atanasova, Jakob~Grue Simonsen, Christina Lioma, and Isabelle Augenstein. 2022.
\newblock \href {https://doi.org/10.1162/tacl_a_00486} {Fact checking with insufficient evidence}.
\newblock \emph{Transactions of the Association for Computational Linguistics}, 10:746--763.

\bibitem[{Atanasova et~al.(2020)Atanasova, Wright, and Augenstein}]{atanasova-etal-2020-generating}
Pepa Atanasova, Dustin Wright, and Isabelle Augenstein. 2020.
\newblock \href {https://doi.org/10.18653/v1/2020.emnlp-main.256} {Generating label cohesive and well-formed adversarial claims}.
\newblock In \emph{Proceedings of the 2020 Conference on Empirical Methods in Natural Language Processing (EMNLP)}, pages 3168--3177.

\bibitem[{Augenstein et~al.(2024)Augenstein, Baldwin, and {Cha et al.}}]{augenstein2024factuality}
Isabelle Augenstein, Timothy Baldwin, and Meeyoung {Cha et al.} 2024.
\newblock \href {https://doi.org/10.1038/s42256-024-00881-z} {Factuality challenges in the era of large language models and opportunities for fact-checking}.
\newblock \emph{Nature Machine Intelligence}, 6:852--863.

\bibitem[{Augenstein et~al.(2019)Augenstein, Lioma, Wang, Chaves~Lima, Hansen, Hansen, and Simonsen}]{augenstein-etal-2019-multifc}
Isabelle Augenstein, Christina Lioma, Dongsheng Wang, Lucas Chaves~Lima, Casper Hansen, Christian Hansen, and Jakob~Grue Simonsen. 2019.
\newblock \href {https://doi.org/10.18653/v1/D19-1475} {{M}ulti{FC}: A real-world multi-domain dataset for evidence-based fact checking of claims}.
\newblock In \emph{Proceedings of the 2019 Conference on Empirical Methods in Natural Language Processing and the 9th International Joint Conference on Natural Language Processing (EMNLP-IJCNLP)}, pages 4685--4697.

\bibitem[{Bekoulis et~al.(2021)Bekoulis, Papagiannopoulou, and Deligiannis}]{bekoulis2021review}
Giannis Bekoulis, Christina Papagiannopoulou, and Nikos Deligiannis. 2021.
\newblock \href {https://doi.org/10.1145/3485127} {A review on fact extraction and verification}.
\newblock \emph{ACM Comput. Surv.}, 55(1).

\bibitem[{Boucher et~al.(2022)Boucher, Shumailov, Anderson, and Papernot}]{boucher2022bad}
Nicholas Boucher, Ilia Shumailov, Ross Anderson, and Nicolas Papernot. 2022.
\newblock \href {https://doi.org/10.1109/SP46214.2022.9833641} {Bad characters: Imperceptible nlp attacks}.
\newblock In \emph{2022 IEEE Symposium on Security and Privacy (SP)}, pages 1987--2004.

\bibitem[{Braun et~al.(2025)Braun, Rothermel, Rohrbach, and Rohrbach}]{braun2025defame}
Tobias Braun, Mark Rothermel, Marcus Rohrbach, and Anna Rohrbach. 2025.
\newblock \href {https://openreview.net/forum?id=umT6rMf1Rm} {{DEFAME}: Dynamic {Evidence}-based {FA}ct-checking with {M}ultimodal {E}xperts}.
\newblock In \emph{Forty-second International Conference on Machine Learning}.

\bibitem[{Cekinel et~al.(2025)Cekinel, Karagoz, and {\c{C}}{\"o}ltekin}]{cekinel-etal-2025-multimodal}
Recep~Firat Cekinel, Pinar Karagoz, and {\c{C}}a{\u{g}}r{\i} {\c{C}}{\"o}ltekin. 2025.
\newblock \href {https://aclanthology.org/2025.coling-main.310/} {Multimodal fact-checking with vision language models: A probing classifier based solution with embedding strategies}.
\newblock In \emph{Proceedings of the 31st International Conference on Computational Linguistics}.

\bibitem[{Chen et~al.(2017)Chen, Fisch, Weston, and Bordes}]{chen-etal-2017-reading}
Danqi Chen, Adam Fisch, Jason Weston, and Antoine Bordes. 2017.
\newblock \href {https://doi.org/10.18653/v1/P17-1171} {Reading {W}ikipedia to answer open-domain questions}.
\newblock In \emph{Proceedings of the 55th Annual Meeting of the Association for Computational Linguistics (Volume 1: Long Papers)}, pages 1870--1879.

\bibitem[{Devlin et~al.(2019)Devlin, Chang, Lee, and Toutanova}]{devlin-etal-2019-bert}
Jacob Devlin, Ming-Wei Chang, Kenton Lee, and Kristina Toutanova. 2019.
\newblock \href {https://doi.org/10.18653/v1/N19-1423} {{BERT}: Pre-training of deep bidirectional transformers for language understanding}.
\newblock In \emph{Proceedings of the 2019 Conference of the North {A}merican Chapter of the Association for Computational Linguistics: Human Language Technologies, Volume 1 (Long and Short Papers)}, pages 4171--4186.

\bibitem[{Du et~al.(2022)Du, Bosselut, and Manning}]{du_2022_synthetic}
Yibing Du, Antoine Bosselut, and Christopher~D. Manning. 2022.
\newblock \href {https://doi.org/10.1609/aaai.v36i10.21302} {Synthetic disinformation attacks on automated fact verification systems}.
\newblock \emph{Proceedings of the AAAI Conference on Artificial Intelligence}, 36(10):10581--10589.

\bibitem[{Eisenschlos et~al.(2021)Eisenschlos, Dhingra, Bulian, B{\"o}rschinger, and Boyd-Graber}]{eisenschlos-etal-2021-fool}
Julian Eisenschlos, Bhuwan Dhingra, Jannis Bulian, Benjamin B{\"o}rschinger, and Jordan Boyd-Graber. 2021.
\newblock \href {https://doi.org/10.18653/v1/2021.naacl-main.32} {Fool me twice: Entailment from {W}ikipedia gamification}.
\newblock In \emph{Proceedings of the 2021 Conference of the North American Chapter of the Association for Computational Linguistics: Human Language Technologies}, pages 352--365.

\bibitem[{Eldifrawi et~al.(2024)Eldifrawi, Wang, and Trabelsi}]{eldifrawi-etal-2024-automated}
Islam Eldifrawi, Shengrui Wang, and Amine Trabelsi. 2024.
\newblock \href {https://doi.org/10.18653/v1/2024.acl-long.361} {Automated justification production for claim veracity in fact checking: A survey on architectures and approaches}.
\newblock In \emph{Proceedings of the 62nd Annual Meeting of the Association for Computational Linguistics (Volume 1: Long Papers)}, pages 6679--6692.

\bibitem[{Gardner et~al.(2018)Gardner, Grus, Neumann, Tafjord, Dasigi, Liu, Peters, Schmitz, and Zettlemoyer}]{gardner-etal-2018-allennlp}
Matt Gardner, Joel Grus, Mark Neumann, Oyvind Tafjord, Pradeep Dasigi, Nelson~F. Liu, Matthew Peters, Michael Schmitz, and Luke Zettlemoyer. 2018.
\newblock \href {https://doi.org/10.18653/v1/W18-2501} {{A}llen{NLP}: A deep semantic natural language processing platform}.
\newblock In \emph{Proceedings of Workshop for {NLP} Open Source Software ({NLP}-{OSS})}, pages 1--6.

\bibitem[{Ghaddar et~al.(2021)Ghaddar, Langlais, Rezagholizadeh, and Rashid}]{ghaddar-etal-2021-end}
Abbas Ghaddar, Phillippe Langlais, Mehdi Rezagholizadeh, and Ahmad Rashid. 2021.
\newblock \href {https://doi.org/10.18653/v1/2021.findings-acl.168} {End-to-end self-debiasing framework for robust {NLU} training}.
\newblock In \emph{Findings of the Association for Computational Linguistics: ACL-IJCNLP 2021}, pages 1923--1929.

\bibitem[{Grattafiori et~al.(2024)Grattafiori, Dubey, Jauhri, and {Pandey et al.}}]{grattafiori2024llama3}
Aaron Grattafiori, Abhimanyu Dubey, Abhinav Jauhri, and Abhinav {Pandey et al.} 2024.
\newblock \href {https://arxiv.org/abs/2407.21783} {The {Llama} 3 herd of models}.
\newblock \emph{Preprint}, arXiv:2407.21783.

\bibitem[{Guo et~al.(2022)Guo, Schlichtkrull, and Vlachos}]{guo-etal-2022-survey}
Zhijiang Guo, Michael Schlichtkrull, and Andreas Vlachos. 2022.
\newblock \href {https://doi.org/10.1162/tacl_a_00454} {A survey on automated fact-checking}.
\newblock \emph{Transactions of the Association for Computational Linguistics}, 10:178--206.

\bibitem[{Gupta and Srikumar(2021)}]{gupta-srikumar-2021-x}
Ashim Gupta and Vivek Srikumar. 2021.
\newblock \href {https://doi.org/10.18653/v1/2021.acl-short.86} {{X}-fact: A new benchmark dataset for multilingual fact checking}.
\newblock In \emph{Proceedings of the 59th Annual Meeting of the Association for Computational Linguistics and the 11th International Joint Conference on Natural Language Processing (Volume 2: Short Papers)}, pages 675--682.

\bibitem[{Hanselowski et~al.(2018)Hanselowski, Zhang, Li, Sorokin, Schiller, Schulz, and Gurevych}]{hanselowski-etal-2018-ukp}
Andreas Hanselowski, Hao Zhang, Zile Li, Daniil Sorokin, Benjamin Schiller, Claudia Schulz, and Iryna Gurevych. 2018.
\newblock \href {https://doi.org/10.18653/v1/W18-5516} {{UKP}-athene: Multi-sentence textual entailment for claim verification}.
\newblock In \emph{Proceedings of the First Workshop on Fact Extraction and {VER}ification ({FEVER})}, pages 103--108.

\bibitem[{He et~al.(2025)He, Li, Zhu, Wen, Cheng, and Lau}]{he2025fact2fiction}
Haorui He, Yupeng Li, Bin~Benjamin Zhu, Dacheng Wen, Reynold Cheng, and Francis C.~M. Lau. 2025.
\newblock \href {https://arxiv.org/abs/2508.06059} {{Fact2Fiction}: Targeted poisoning attack to agentic fact-checking system}.
\newblock \emph{Preprint}, arXiv:2508.06059.

\bibitem[{He et~al.(2023)He, Gao, and Chen}]{he2023debertav3}
Pengcheng He, Jianfeng Gao, and Weizhu Chen. 2023.
\newblock \href {https://openreview.net/forum?id=sE7-XhLxHA} {{DeBERTaV3}: Improving deberta using electra-style pre-training with gradient-disentangled embedding sharing}.
\newblock In \emph{International Conference on Learning Representations}.

\bibitem[{He et~al.(2021)He, Liu, Gao, and Chen}]{he2021deberta}
Pengcheng He, Xiaodong Liu, Jianfeng Gao, and Weizhu Chen. 2021.
\newblock \href {https://openreview.net/forum?id=XPZIaotutsD} {{DeBERTa}: Decoding-enhanced bert with disentangled attention}.
\newblock In \emph{International Conference on Learning Representations}.

\bibitem[{Hidey et~al.(2020)Hidey, Chakrabarty, Alhindi, Varia, Krstovski, Diab, and Muresan}]{hidey-etal-2020-deseption}
Christopher Hidey, Tuhin Chakrabarty, Tariq Alhindi, Siddharth Varia, Kriste Krstovski, Mona Diab, and Smaranda Muresan. 2020.
\newblock \href {https://doi.org/10.18653/v1/2020.acl-main.761} {{D}e{S}e{P}tion: Dual sequence prediction and adversarial examples for improved fact-checking}.
\newblock In \emph{Proceedings of the 58th Annual Meeting of the Association for Computational Linguistics}, pages 8593--8606.

\bibitem[{Hidey and Diab(2018)}]{hidey-diab-2018-team}
Christopher Hidey and Mona Diab. 2018.
\newblock \href {https://doi.org/10.18653/v1/W18-5525} {Team {SWEEP}er: Joint sentence extraction and fact checking with pointer networks}.
\newblock In \emph{Proceedings of the First Workshop on Fact Extraction and {VER}ification ({FEVER})}, pages 150--155.

\bibitem[{Hinton(2002)}]{hinton2002}
Geoffrey~E. Hinton. 2002.
\newblock \href {https://doi.org/10.1162/089976602760128018} {Training products of experts by minimizing contrastive divergence}.
\newblock \emph{Neural Computation}, 14(8):1771--1800.

\bibitem[{Hu et~al.(2025)Hu, Long, and Wang}]{hu-etal-2025-decomposition}
Qisheng Hu, Quanyu Long, and Wenya Wang. 2025.
\newblock \href {https://doi.org/10.18653/v1/2025.naacl-long.320} {Decomposition dilemmas: Does claim decomposition boost or burden fact-checking performance?}
\newblock In \emph{Proceedings of the 2025 Conference of the Nations of the Americas Chapter of the Association for Computational Linguistics: Human Language Technologies (Volume 1: Long Papers)}, pages 6313--6336.

\bibitem[{Hu et~al.(2022)Hu, Guo, Wu, Liu, Wen, and Yu}]{hu-etal-2022-chef}
Xuming Hu, Zhijiang Guo, GuanYu Wu, Aiwei Liu, Lijie Wen, and Philip Yu. 2022.
\newblock \href {https://doi.org/10.18653/v1/2022.naacl-main.246} {{CHEF}: A pilot {C}hinese dataset for evidence-based fact-checking}.
\newblock In \emph{Proceedings of the 2022 Conference of the North American Chapter of the Association for Computational Linguistics: Human Language Technologies}, pages 3362--3376.

\bibitem[{Jiang et~al.(2023)Jiang, Sablayrolles, Mensch, Bamford, Chaplot, de~las Casas, Bressand, Lengyel, Lample, Saulnier, Lavaud, Lachaux, Stock, Scao, Lavril, Wang, Lacroix, and Sayed}]{jiang2023mistral7b}
Albert~Q. Jiang, Alexandre Sablayrolles, Arthur Mensch, Chris Bamford, Devendra~Singh Chaplot, Diego de~las Casas, Florian Bressand, Gianna Lengyel, Guillaume Lample, Lucile Saulnier, Lélio~Renard Lavaud, Marie-Anne Lachaux, Pierre Stock, Teven~Le Scao, Thibaut Lavril, Thomas Wang, Timothée Lacroix, and William~El Sayed. 2023.
\newblock \href {https://arxiv.org/abs/2310.06825} {Mistral 7b}.
\newblock \emph{Preprint}, arXiv:2310.06825.

\bibitem[{Jiang et~al.(2020)Jiang, Bordia, Zhong, Dognin, Singh, and Bansal}]{jiang-etal-2020-hover}
Yichen Jiang, Shikha Bordia, Zheng Zhong, Charles Dognin, Maneesh Singh, and Mohit Bansal. 2020.
\newblock \href {https://doi.org/10.18653/v1/2020.findings-emnlp.309} {{H}o{V}er: A dataset for many-hop fact extraction and claim verification}.
\newblock In \emph{Findings of the Association for Computational Linguistics: EMNLP 2020}, pages 3441--3460.

\bibitem[{Jin et~al.(2020)Jin, Jin, Zhou, and Szolovits}]{jin2020is}
Di~Jin, Zhijing Jin, Joey~Tianyi Zhou, and Peter Szolovits. 2020.
\newblock \href {https://doi.org/10.1609/aaai.v34i05.6311} {Is {BERT} really robust? a strong baseline for natural language attack on text classification and entailment}.
\newblock \emph{Proceedings of the AAAI Conference on Artificial Intelligence}, 34(05):8018--8025.

\bibitem[{Karimi~Mahabadi et~al.(2020)Karimi~Mahabadi, Belinkov, and Henderson}]{karimi-mahabadi-etal-2020-end}
Rabeeh Karimi~Mahabadi, Yonatan Belinkov, and James Henderson. 2020.
\newblock \href {https://doi.org/10.18653/v1/2020.acl-main.769} {End-to-end bias mitigation by modelling biases in corpora}.
\newblock In \emph{Proceedings of the 58th Annual Meeting of the Association for Computational Linguistics}, pages 8706--8716.

\bibitem[{Karpukhin et~al.(2020)Karpukhin, Oguz, Min, Lewis, Wu, Edunov, Chen, and Yih}]{karpukhin-etal-2020-dense}
Vladimir Karpukhin, Barlas Oguz, Sewon Min, Patrick Lewis, Ledell Wu, Sergey Edunov, Danqi Chen, and Wen-tau Yih. 2020.
\newblock \href {https://doi.org/10.18653/v1/2020.emnlp-main.550} {Dense passage retrieval for open-domain question answering}.
\newblock In \emph{Proceedings of the 2020 Conference on Empirical Methods in Natural Language Processing (EMNLP)}, pages 6769--6781.

\bibitem[{Kim et~al.(2021)Kim, Kim, Hong, and Kim}]{kim-etal-2021-robust}
Byeongchang Kim, Hyunwoo Kim, Seokhee Hong, and Gunhee Kim. 2021.
\newblock \href {https://doi.org/10.18653/v1/2021.naacl-main.121} {How robust are fact checking systems on colloquial claims?}
\newblock In \emph{Proceedings of the 2021 Conference of the North American Chapter of the Association for Computational Linguistics: Human Language Technologies}, pages 1535--1548.

\bibitem[{Kim and Allan(2019)}]{kim-allan-2019-fever}
Youngwoo Kim and James Allan. 2019.
\newblock \href {https://doi.org/10.18653/v1/D19-6615} {{FEVER} breaker{'}s run of team {N}b{A}uz{D}r{L}qg}.
\newblock In \emph{Proceedings of the Second Workshop on Fact Extraction and VERification (FEVER)}, pages 99--104.

\bibitem[{Kotonya and Toni(2020)}]{kotonya-toni-2020-explainable}
Neema Kotonya and Francesca Toni. 2020.
\newblock \href {https://doi.org/10.18653/v1/2020.coling-main.474} {Explainable automated fact-checking: A survey}.
\newblock In \emph{Proceedings of the 28th International Conference on Computational Linguistics}, pages 5430--5443.

\bibitem[{Kruengkrai et~al.(2021)Kruengkrai, Yamagishi, and Wang}]{kruengkrai-etal-2021-multi}
Canasai Kruengkrai, Junichi Yamagishi, and Xin Wang. 2021.
\newblock \href {https://doi.org/10.18653/v1/2021.findings-acl.217} {A multi-level attention model for evidence-based fact checking}.
\newblock In \emph{Findings of the Association for Computational Linguistics: ACL-IJCNLP 2021}, pages 2447--2460.

\bibitem[{Kwiatkowski et~al.(2019)Kwiatkowski, Palomaki, Redfield, Collins, Parikh, Alberti, Epstein, Polosukhin, Devlin, Lee, Toutanova, Jones, Kelcey, Chang, Dai, Uszkoreit, Le, and Petrov}]{kwiatkowski-etal-2019-natural}
Tom Kwiatkowski, Jennimaria Palomaki, Olivia Redfield, Michael Collins, Ankur Parikh, Chris Alberti, Danielle Epstein, Illia Polosukhin, Jacob Devlin, Kenton Lee, Kristina Toutanova, Llion Jones, Matthew Kelcey, Ming-Wei Chang, Andrew~M. Dai, Jakob Uszkoreit, Quoc Le, and Slav Petrov. 2019.
\newblock \href {https://doi.org/10.1162/tacl_a_00276} {Natural questions: A benchmark for question answering research}.
\newblock \emph{Transactions of the Association for Computational Linguistics}, 7:452--466.

\bibitem[{Lan et~al.(2020)Lan, Chen, Goodman, Gimpel, Sharma, and Soricut}]{lan2020ALBERT}
Zhenzhong Lan, Mingda Chen, Sebastian Goodman, Kevin Gimpel, Piyush Sharma, and Radu Soricut. 2020.
\newblock \href {https://openreview.net/forum?id=H1eA7AEtvS} {{ALBERT}: A lite bert for self-supervised learning of language representations}.
\newblock In \emph{International Conference on Learning Representations}.

\bibitem[{Lee et~al.(2021)Lee, Won, Kim, Lee, Park, and Jung}]{lww2021crossaug}
Minwoo Lee, Seungpil Won, Juae Kim, Hwanhee Lee, Cheoneum Park, and Kyomin Jung. 2021.
\newblock \href {https://doi.org/10.1145/3459637.3482078} {{CrossAug}: A contrastive data augmentation method for debiasing fact verification models}.
\newblock In \emph{Proceedings of the 30th ACM International Conference on Information \& Knowledge Management}, page 3181–3185.

\bibitem[{Lewis et~al.(2020)Lewis, Liu, Goyal, Ghazvininejad, Mohamed, Levy, Stoyanov, and Zettlemoyer}]{lewis-etal-2020-bart}
Mike Lewis, Yinhan Liu, Naman Goyal, Marjan Ghazvininejad, Abdelrahman Mohamed, Omer Levy, Veselin Stoyanov, and Luke Zettlemoyer. 2020.
\newblock \href {https://doi.org/10.18653/v1/2020.acl-main.703} {{BART}: Denoising sequence-to-sequence pre-training for natural language generation, translation, and comprehension}.
\newblock In \emph{Proceedings of the 58th Annual Meeting of the Association for Computational Linguistics}, pages 7871--7880.

\bibitem[{Li et~al.(2020)Li, Ma, Guo, Xue, and Qiu}]{li-etal-2020-bert-attack}
Linyang Li, Ruotian Ma, Qipeng Guo, Xiangyang Xue, and Xipeng Qiu. 2020.
\newblock \href {https://doi.org/10.18653/v1/2020.emnlp-main.500} {{BERT}-{ATTACK}: Adversarial attack against {BERT} using {BERT}}.
\newblock In \emph{Proceedings of the 2020 Conference on Empirical Methods in Natural Language Processing (EMNLP)}, pages 6193--6202.

\bibitem[{Liu et~al.(2019)Liu, Ott, Goyal, Du, Joshi, Chen, Levy, Lewis, Zettlemoyer, and Stoyanov}]{liu2019roberta}
Yinhan Liu, Myle Ott, Naman Goyal, Jingfei Du, Mandar Joshi, Danqi Chen, Omer Levy, Mike Lewis, Luke Zettlemoyer, and Veselin Stoyanov. 2019.
\newblock \href {https://arxiv.org/abs/1907.11692} {{RoBERTa}: A robustly optimized {BERT} pretraining approach}.
\newblock \emph{Preprint}, arXiv:1907.11692.

\bibitem[{Liu et~al.(2020)Liu, Xiong, Sun, and Liu}]{liu-etal-2020-fine}
Zhenghao Liu, Chenyan Xiong, Maosong Sun, and Zhiyuan Liu. 2020.
\newblock \href {https://doi.org/10.18653/v1/2020.acl-main.655} {Fine-grained fact verification with kernel graph attention network}.
\newblock In \emph{Proceedings of the 58th Annual Meeting of the Association for Computational Linguistics}, pages 7342--7351.

\bibitem[{Malon(2018)}]{malon-2018-team}
Christopher Malon. 2018.
\newblock \href {https://doi.org/10.18653/v1/W18-5517} {Team papelo: Transformer networks at {FEVER}}.
\newblock In \emph{Proceedings of the First Workshop on Fact Extraction and {VER}ification ({FEVER})}, pages 109--113.

\bibitem[{Mamta and Cocarascu(2025)}]{factual2025mamta}
Mamta Mamta and Oana Cocarascu. 2025.
\newblock \href {https://aclanthology.org/2025.naacl-long.534/} {{F}act{E}val: Evaluating the robustness of fact verification systems in the era of large language models}.
\newblock In \emph{Proceedings of the 2025 Conference of the Nations of the Americas Chapter of the Association for Computational Linguistics: Human Language Technologies (Volume 1: Long Papers)}, pages 10647--10660.

\bibitem[{Martel and Rand(2024)}]{martel2024nature}
Cameron Martel and David~G. Rand. 2024.
\newblock \href {https://doi.org/10.1162/tacl_a_00454} {Fact-checker warning labels are effective even for those who distrust fact-checkers}.
\newblock \emph{Nature Human Behaviour}, 8:1957--1967.

\bibitem[{Mesnard et~al.(2024)Mesnard, Hardin, and {Dadashi et al.}}]{gemmateam2024}
Thomas Mesnard, Cassidy Hardin, and Robert {Dadashi et al.} 2024.
\newblock \href {https://arxiv.org/abs/2403.08295} {Gemma: Open models based on gemini research and technology}.
\newblock \emph{Preprint}, arXiv:2403.08295.

\bibitem[{Nguyen et~al.(2016)Nguyen, Rosenberg, Song, Gao, Tiwary, Majumder, and Deng}]{nguyen2016msmarco}
Tri Nguyen, Mir Rosenberg, Xia Song, Jianfeng Gao, Saurabh Tiwary, Rangan Majumder, and Li~Deng. 2016.
\newblock \href {https://arxiv.org/abs/1611.09268v2} {Ms marco: A human generated machine reading comprehension dataset}.
\newblock \emph{Preprint}, arXiv:1611.09268v2.

\bibitem[{Nie et~al.(2019)Nie, Chen, and Bansal}]{Nie_Chen_Bansal_2019}
Yixin Nie, Haonan Chen, and Mohit Bansal. 2019.
\newblock \href {https://doi.org/10.1609/aaai.v33i01.33016859} {Combining fact extraction and verification with neural semantic matching networks}.
\newblock \emph{Proceedings of the AAAI Conference on Artificial Intelligence}, 33(01):6859--6866.

\bibitem[{Niewinski et~al.(2019)Niewinski, Pszona, and Janicka}]{niewinski-etal-2019-gem}
Piotr Niewinski, Maria Pszona, and Maria Janicka. 2019.
\newblock \href {https://doi.org/10.18653/v1/D19-6604} {{GEM}: Generative enhanced model for adversarial attacks}.
\newblock In \emph{Proceedings of the Second Workshop on Fact Extraction and VERification (FEVER)}, pages 20--26.

\bibitem[{OpenAI(2022)}]{openai2022gpt3-5}
OpenAI. 2022.
\newblock Chatgpt {Blog} {Post}.
\newblock \url{https://openai.com/index/chatgpt/}.

\bibitem[{OpenAI(2024)}]{openai2024gpt4technicalreport}
OpenAI. 2024.
\newblock \href {https://arxiv.org/abs/2303.08774} {{GPT-4} {Technical} {Report}}.
\newblock \emph{Preprint}, arXiv:2303.08774.

\bibitem[{Ostrowski et~al.(2021)Ostrowski, Arora, Atanasova, and Augenstein}]{ostrowski2021multi}
Wojciech Ostrowski, Arnav Arora, Pepa Atanasova, and Isabelle Augenstein. 2021.
\newblock \href {https://doi.org/10.24963/ijcai.2021/536} {Multi-hop fact checking of political claims}.
\newblock In \emph{Proceedings of the Thirtieth International Joint Conference on Artificial Intelligence}, pages 3892--3898.

\bibitem[{Parikh et~al.(2016)Parikh, T{\"a}ckstr{\"o}m, Das, and Uszkoreit}]{parikh-etal-2016-decomposable}
Ankur Parikh, Oscar T{\"a}ckstr{\"o}m, Dipanjan Das, and Jakob Uszkoreit. 2016.
\newblock \href {https://doi.org/10.18653/v1/D16-1244} {A decomposable attention model for natural language inference}.
\newblock In \emph{Proceedings of the 2016 Conference on Empirical Methods in Natural Language Processing}, pages 2249--2255.

\bibitem[{Pradeep et~al.(2021)Pradeep, Ma, Nogueira, and Lin}]{pradeep-etal-2021-scientific}
Ronak Pradeep, Xueguang Ma, Rodrigo Nogueira, and Jimmy Lin. 2021.
\newblock \href {https://aclanthology.org/2021.louhi-1.11} {Scientific claim verification with {V}er{T}5erini}.
\newblock In \emph{Proceedings of the 12th International Workshop on Health Text Mining and Information Analysis}, pages 94--103.

\bibitem[{Przyby{\l}a et~al.(2024)Przyby{\l}a, Wu, Shvets, Mu, Sheang, Song, and Saggion}]{clef-checkthat:2024:task6}
Piotr Przyby{\l}a, Ben Wu, Alexander Shvets, Yida Mu, Kim~Cheng Sheang, Xingyi Song, and Horacio Saggion. 2024.
\newblock Overview of the {CLEF}-2024 {CheckThat}! lab task 6 on robustness of credibility assessment with adversarial examples (incrediblae).
\newblock In \emph{Working Notes of CLEF 2024 - Conference and Labs of the Evaluation Forum}, CLEF~2024.

\bibitem[{Przybyła(2024)}]{przybyła2024attacking}
Piotr Przybyła. 2024.
\newblock \href {https://arxiv.org/abs/2410.20940} {Attacking misinformation detection using adversarial examples generated by language models}.
\newblock \emph{Preprint}, arXiv:2410.20940.

\bibitem[{Radford et~al.(2019)Radford, Wu, Child, Luan, Amodei, and Sutskever}]{radford2019language}
Alec Radford, Jeff Wu, Rewon Child, David Luan, Dario Amodei, and Ilya Sutskever. 2019.
\newblock \href {https://cdn.openai.com/better-language-models/language_models_are_unsupervised_multitask_learners.pdf} {Language models are unsupervised multitask learners}.
\newblock \emph{OpenAI Blog}, 1(8):9.

\bibitem[{Raffel et~al.(2020)Raffel, Shazeer, Roberts, Lee, Narang, Matena, Zhou, Li, and Liu}]{raffel2020exploring}
Colin Raffel, Noam Shazeer, Adam Roberts, Katherine Lee, Sharan Narang, Michael Matena, Yanqi Zhou, Wei Li, and Peter~J. Liu. 2020.
\newblock \href {http://jmlr.org/papers/v21/20-074.html} {Exploring the limits of transfer learning with a unified text-to-text transformer}.
\newblock \emph{Journal of Machine Learning Research}, 21(140):1--67.

\bibitem[{Ribeiro et~al.(2018)Ribeiro, Singh, and Guestrin}]{ribeiro-etal-2018-semantically}
Marco~Tulio Ribeiro, Sameer Singh, and Carlos Guestrin. 2018.
\newblock \href {https://doi.org/10.18653/v1/P18-1079} {Semantically equivalent adversarial rules for debugging {NLP} models}.
\newblock In \emph{Proceedings of the 56th Annual Meeting of the Association for Computational Linguistics (Volume 1: Long Papers)}, pages 856--865.

\bibitem[{Robertson et~al.(1994)Robertson, Walker, Jones, Hancock{-}Beaulieu, and Gatford}]{robertson1994okapi}
Stephen~E. Robertson, Steve Walker, Susan Jones, Micheline Hancock{-}Beaulieu, and Mike Gatford. 1994.
\newblock \href {http://trec.nist.gov/pubs/trec3/papers/city.ps.gz} {Okapi at {TREC-3}}.
\newblock In \emph{Proceedings of The Third Text REtrieval Conference, {TREC}}, volume 500-225 of \emph{{NIST} Special Publication}, pages 109--126.

\bibitem[{Rothermel et~al.(2024)Rothermel, Braun, Rohrbach, and Rohrbach}]{rothermel-etal-2024-infact}
Mark Rothermel, Tobias Braun, Marcus Rohrbach, and Anna Rohrbach. 2024.
\newblock \href {https://doi.org/10.18653/v1/2024.fever-1.12} {{I}n{F}act: A strong baseline for automated fact-checking}.
\newblock In \emph{Proceedings of the Seventh Fact Extraction and VERification Workshop (FEVER)}, pages 108--112.

\bibitem[{Saakyan et~al.(2021)Saakyan, Chakrabarty, and Muresan}]{saakyan-etal-2021-covid}
Arkadiy Saakyan, Tuhin Chakrabarty, and Smaranda Muresan. 2021.
\newblock \href {https://doi.org/10.18653/v1/2021.acl-long.165} {{COVID}-{F}act: Fact extraction and verification of real-world claims on {COVID}-19 pandemic}.
\newblock In \emph{Proceedings of the 59th Annual Meeting of the Association for Computational Linguistics and the 11th International Joint Conference on Natural Language Processing (Volume 1: Long Papers)}, pages 2116--2129.

\bibitem[{Sai et~al.(2021)Sai, Dixit, Sheth, Mohan, and Khapra}]{sai-etal-2021-perturbation}
Ananya~B. Sai, Tanay Dixit, Dev~Yashpal Sheth, Sreyas Mohan, and Mitesh~M. Khapra. 2021.
\newblock \href {https://doi.org/10.18653/v1/2021.emnlp-main.575} {Perturbation {C}heck{L}ists for evaluating {NLG} evaluation metrics}.
\newblock In \emph{Proceedings of the 2021 Conference on Empirical Methods in Natural Language Processing}, pages 7219--7234.

\bibitem[{Sakib et~al.(2025)Sakib, Das, and Ahmed}]{sakib-etal-2025-battling}
Shahnewaz~Karim Sakib, Anindya~Bijoy Das, and Shibbir Ahmed. 2025.
\newblock \href {https://doi.org/10.18653/v1/2025.trustnlp-main.28} {Battling misinformation: An empirical study on adversarial factuality in open-source large language models}.
\newblock In \emph{Proceedings of the 5th Workshop on Trustworthy NLP (TrustNLP 2025)}, pages 432--443.

\bibitem[{Samarinas et~al.(2020)Samarinas, Hsu, and Lee}]{samarinas2020latent}
Chris Samarinas, Wynne Hsu, and Mong~Li Lee. 2020.
\newblock \href {https://doi.org/10.1109/ICTAI50040.2020.00147} {Latent retrieval for large-scale fact-checking and question answering with nli training}.
\newblock In \emph{2020 IEEE 32nd International Conference on Tools with Artificial Intelligence (ICTAI)}, pages 941--948.

\bibitem[{Samarinas et~al.(2021)Samarinas, Hsu, and Lee}]{samarinas2021improving}
Chris Samarinas, Wynne Hsu, and Mong~Li Lee. 2021.
\newblock \href {https://doi.org/10.18653/v1/2021.naacl-demos.10} {Improving evidence retrieval for automated explainable fact-checking}.
\newblock In \emph{Proceedings of the 2021 Conference of the North American Chapter of the Association for Computational Linguistics: Human Language Technologies: Demonstrations}, pages 84--91.

\bibitem[{Schlichtkrull et~al.(2023)Schlichtkrull, Guo, and Vlachos}]{schlichtkrull2023AVeriTeC}
Michael Schlichtkrull, Zhijiang Guo, and Andreas Vlachos. 2023.
\newblock \href {https://proceedings.neurips.cc/paper_files/paper/2023/file/cd86a30526cd1aff61d6f89f107634e4-Paper-Datasets_and_Benchmarks.pdf} {{AVeriTeC}: A dataset for real-world claim verification with evidence from the web}.
\newblock In \emph{Advances in Neural Information Processing Systems}, volume~36, pages 65128--65167.

\bibitem[{Schuster et~al.(2021)Schuster, Fisch, and Barzilay}]{schuster-etal-2021-get}
Tal Schuster, Adam Fisch, and Regina Barzilay. 2021.
\newblock \href {https://doi.org/10.18653/v1/2021.naacl-main.52} {Get your vitamin {C}! robust fact verification with contrastive evidence}.
\newblock In \emph{Proceedings of the 2021 Conference of the North American Chapter of the Association for Computational Linguistics: Human Language Technologies}, pages 624--643.

\bibitem[{Schuster et~al.(2019)Schuster, Shah, Yeo, Roberto Filizzola~Ortiz, Santus, and Barzilay}]{schuster2019towards}
Tal Schuster, Darsh Shah, Yun Jie~Serene Yeo, Daniel Roberto Filizzola~Ortiz, Enrico Santus, and Regina Barzilay. 2019.
\newblock \href {https://doi.org/10.18653/v1/D19-1341} {Towards debiasing fact verification models}.
\newblock In \emph{Proceedings of the 2019 Conference on Empirical Methods in Natural Language Processing and the 9th International Joint Conference on Natural Language Processing (EMNLP-IJCNLP)}, pages 3419--3425.

\bibitem[{Thorne et~al.(2018{\natexlab{a}})Thorne, Vlachos, Christodoulopoulos, and Mittal}]{thorne-etal-2018-fever}
James Thorne, Andreas Vlachos, Christos Christodoulopoulos, and Arpit Mittal. 2018{\natexlab{a}}.
\newblock \href {https://doi.org/10.18653/v1/N18-1074} {{FEVER}: a large-scale dataset for fact extraction and {VER}ification}.
\newblock In \emph{Proceedings of the 2018 Conference of the North {A}merican Chapter of the Association for Computational Linguistics: Human Language Technologies, Volume 1 (Long Papers)}, pages 809--819.

\bibitem[{Thorne et~al.(2019{\natexlab{a}})Thorne, Vlachos, Christodoulopoulos, and Mittal}]{thorne-etal-2019-evaluating}
James Thorne, Andreas Vlachos, Christos Christodoulopoulos, and Arpit Mittal. 2019{\natexlab{a}}.
\newblock \href {https://doi.org/10.18653/v1/D19-1292} {Evaluating adversarial attacks against multiple fact verification systems}.
\newblock In \emph{Proceedings of the 2019 Conference on Empirical Methods in Natural Language Processing and the 9th International Joint Conference on Natural Language Processing (EMNLP-IJCNLP)}, pages 2944--2953.

\bibitem[{Thorne et~al.(2018{\natexlab{b}})Thorne, Vlachos, Cocarascu, Christodoulopoulos, and Mittal}]{thorne-etal-2018-fact}
James Thorne, Andreas Vlachos, Oana Cocarascu, Christos Christodoulopoulos, and Arpit Mittal. 2018{\natexlab{b}}.
\newblock \href {https://doi.org/10.18653/v1/W18-5501} {The fact extraction and {VER}ification ({FEVER}) shared task}.
\newblock In \emph{Proceedings of the First Workshop on Fact Extraction and {VER}ification ({FEVER})}, pages 1--9.

\bibitem[{Thorne et~al.(2019{\natexlab{b}})Thorne, Vlachos, Cocarascu, Christodoulopoulos, and Mittal}]{thorne-etal-2019-fever2}
James Thorne, Andreas Vlachos, Oana Cocarascu, Christos Christodoulopoulos, and Arpit Mittal. 2019{\natexlab{b}}.
\newblock \href {https://doi.org/10.18653/v1/D19-6601} {The {FEVER}2.0 shared task}.
\newblock In \emph{Proceedings of the Second Workshop on Fact Extraction and VERification (FEVER)}, pages 1--6.

\bibitem[{Tian et~al.(2022)Tian, Cao, Zhang, and Xing}]{tian2022debiasing}
Bing Tian, Yixin Cao, Yong Zhang, and Chunxiao Xing. 2022.
\newblock \href {https://ojs.aaai.org/index.php/AAAI/article/view/21389} {Debiasing {NLU} models via causal intervention and counterfactual reasoning}.
\newblock \emph{Proceedings of the AAAI Conference on Artificial Intelligence}, 36(10):11376--11384.

\bibitem[{Vladika and Matthes(2023)}]{vladika-matthes-2023-scientific}
Juraj Vladika and Florian Matthes. 2023.
\newblock \href {https://doi.org/10.18653/v1/2023.findings-acl.387} {Scientific fact-checking: A survey of resources and approaches}.
\newblock In \emph{Findings of the Association for Computational Linguistics: ACL 2023}, pages 6215--6230.

\bibitem[{Vykopal et~al.(2024)Vykopal, Pikuliak, Ostermann, and Šimko}]{vykopal2024generative}
Ivan Vykopal, Matúš Pikuliak, Simon Ostermann, and Marián Šimko. 2024.
\newblock \href {https://arxiv.org/abs/2407.02351} {Generative large language models in automated fact-checking: A survey}.
\newblock \emph{Preprint}, arXiv:2407.02351.

\bibitem[{Wadden et~al.(2020)Wadden, Lin, Lo, Wang, van Zuylen, Cohan, and Hajishirzi}]{wadden-etal-2020-fact}
David Wadden, Shanchuan Lin, Kyle Lo, Lucy~Lu Wang, Madeleine van Zuylen, Arman Cohan, and Hannaneh Hajishirzi. 2020.
\newblock \href {https://doi.org/10.18653/v1/2020.emnlp-main.609} {Fact or fiction: Verifying scientific claims}.
\newblock In \emph{Proceedings of the 2020 Conference on Empirical Methods in Natural Language Processing (EMNLP)}, pages 7534--7550.

\bibitem[{Walter et~al.(2020)Walter, Cohen, Holbert, and Morag}]{walter2020fact}
Nathan Walter, Jonathan Cohen, R.~Lance Holbert, and Yasmin Morag. 2020.
\newblock \href {https://doi.org/10.1080/10584609.2019.1668894} {Fact-checking: A meta-analysis of what works and for whom}.
\newblock \emph{Political Communication}, 37(3):350--375.

\bibitem[{Wang et~al.(2023)Wang, Dou, Chen, Sun, Yu, and Shu}]{wang2023attacking}
Haoran Wang, Yingtong Dou, Canyu Chen, Lichao Sun, Philip~S. Yu, and Kai Shu. 2023.
\newblock \href {https://doi.org/10.1145/3543507.3583868} {Attacking fake news detectors via manipulating news social engagement}.
\newblock In \emph{Proceedings of the ACM Web Conference 2023}, page 3978–3986.

\bibitem[{Wu et~al.(2019)Wu, Morstatter, Carley, and Liu}]{wu2019misinfo}
Liang Wu, Fred Morstatter, Kathleen~M. Carley, and Huan Liu. 2019.
\newblock \href {https://doi.org/10.1145/3373464.3373475} {Misinformation in social media: Definition, manipulation, and detection}.
\newblock \emph{SIGKDD Explor. Newsl.}, 21(2):80–90.

\bibitem[{Xu et~al.(2023)Xu, Liu, Wu, and Wang}]{xu2023counterfactual}
Weizhi Xu, Qiang Liu, Shu Wu, and Liang Wang. 2023.
\newblock \href {https://doi.org/10.18653/v1/2023.acl-long.374} {Counterfactual debiasing for fact verification}.
\newblock In \emph{Proceedings of the 61st Annual Meeting of the Association for Computational Linguistics (Volume 1: Long Papers)}, pages 6777--6789.

\bibitem[{Yang et~al.(2024)Yang, Christensen, Gilda, Fernandes, Oliveira, Wilson, and Woodard}]{yang2024factchecking}
Qianfeng Yang, Tess Christensen, Shlok Gilda, Juliana Fernandes, Daniela Oliveira, Ronald Wilson, and Damon Woodard. 2024.
\newblock \href {https://arxiv.org/abs/2402.13244} {Are fact-checking tools helpful? an exploration of the usability of google fact check}.
\newblock \emph{Preprint}, arXiv:2402.13244.

\bibitem[{Ye et~al.(2020)Ye, Lin, Du, Liu, Li, Sun, and Liu}]{ye-etal-2020-coreferential}
Deming Ye, Yankai Lin, Jiaju Du, Zhenghao Liu, Peng Li, Maosong Sun, and Zhiyuan Liu. 2020.
\newblock \href {https://doi.org/10.18653/v1/2020.emnlp-main.582} {{C}oreferential {R}easoning {L}earning for {L}anguage {R}epresentation}.
\newblock In \emph{Proceedings of the 2020 Conference on Empirical Methods in Natural Language Processing (EMNLP)}, pages 7170--7186.

\bibitem[{Yoneda et~al.(2018)Yoneda, Mitchell, Welbl, Stenetorp, and Riedel}]{yoneda-etal-2018-ucl}
Takuma Yoneda, Jeff Mitchell, Johannes Welbl, Pontus Stenetorp, and Sebastian Riedel. 2018.
\newblock \href {https://doi.org/10.18653/v1/W18-5515} {{UCL} machine reading group: Four factor framework for fact finding ({H}exa{F})}.
\newblock In \emph{Proceedings of the First Workshop on Fact Extraction and {VER}ification ({FEVER})}, pages 97--102.

\bibitem[{Zellers et~al.(2019)Zellers, Holtzman, Rashkin, Bisk, Farhadi, Roesner, and Choi}]{zellers2019defending}
Rowan Zellers, Ari Holtzman, Hannah Rashkin, Yonatan Bisk, Ali Farhadi, Franziska Roesner, and Yejin Choi. 2019.
\newblock \href {https://proceedings.neurips.cc/paper_files/paper/2019/file/3e9f0fc9b2f89e043bc6233994dfcf76-Paper.pdf} {Defending against neural fake news}.
\newblock In \emph{Advances in Neural Information Processing Systems}, volume~32.

\bibitem[{Zeng and Gao(2024)}]{zeng-etal-tacl-2024}
Fengzhu Zeng and Wei Gao. 2024.
\newblock \href {https://doi.org/10.1162/tacl_a_00649} {{JustiLM}: Few-shot justification generation for explainable fact-checking of real-world claims}.
\newblock \emph{Transactions of the Association for Computational Linguistics}, 12:334--354.

\bibitem[{Zhang et~al.(2024{\natexlab{a}})Zhang, Guo, and Vlachos}]{zhang2024need}
Caiqi Zhang, Zhijiang Guo, and Andreas Vlachos. 2024{\natexlab{a}}.
\newblock \href {https://doi.org/10.18653/v1/2024.emnlp-main.113} {Do we need language-specific fact-checking models? the case of {C}hinese}.
\newblock In \emph{Proceedings of the 2024 Conference on Empirical Methods in Natural Language Processing}, pages 1899--1914.

\bibitem[{Zhang et~al.(2024{\natexlab{b}})Zhang, Zhang, and Zhou}]{zhang2024causal}
Congzhi Zhang, Linhai Zhang, and Deyu Zhou. 2024{\natexlab{b}}.
\newblock \href {https://doi.org/10.1609/aaai.v38i17.29925} {Causal walk: Debiasing multi-hop fact verification with front-door adjustment}.
\newblock \emph{Proceedings of the AAAI Conference on Artificial Intelligence}, 38(17):19533--19541.

\bibitem[{Zhang et~al.(2024{\natexlab{c}})Zhang, Liu, Xie, Zhang, Xu, and Zha}]{zhang2024escnet}
Fanrui Zhang, Jiawei Liu, Jingyi Xie, Qiang Zhang, Yongchao Xu, and Zheng-Jun Zha. 2024{\natexlab{c}}.
\newblock \href {https://doi.org/10.1145/3589334.3645455} {{ESCNet}: Entity-enhanced and stance checking network for multi-modal fact-checking}.
\newblock In \emph{Proceedings of the ACM Web Conference 2024}, page 2429–2440.

\bibitem[{Zhang et~al.(2023)Zhang, Gan, Wang, Zhang, Zhang, Yang, Gao, Wu, Dong, He, Zhuo, Yang, Huang, Li, Wu, Lu, Zhu, Chen, Han, Pan, Wang, Wang, Wu, Zeng, and Chen}]{zhang2023fengshenbang}
Jiaxing Zhang, Ruyi Gan, Junjie Wang, Yuxiang Zhang, Lin Zhang, Ping Yang, Xinyu Gao, Ziwei Wu, Xiaoqun Dong, Junqing He, Jianheng Zhuo, Qi~Yang, Yongfeng Huang, Xiayu Li, Yanghan Wu, Junyu Lu, Xinyu Zhu, Weifeng Chen, Ting Han, Kunhao Pan, Rui Wang, Hao Wang, Xiaojun Wu, Zhongshen Zeng, and Chongpei Chen. 2023.
\newblock \href {https://arxiv.org/abs/2209.02970} {Fengshenbang 1.0: Being the foundation of chinese cognitive intelligence}.
\newblock \emph{Preprint}, arXiv:2209.02970.

\bibitem[{Zhang et~al.(2020{\natexlab{a}})Zhang, Zhao, Saleh, and Liu}]{pmlr-v119-zhang20ae}
Jingqing Zhang, Yao Zhao, Mohammad Saleh, and Peter Liu. 2020{\natexlab{a}}.
\newblock \href {https://proceedings.mlr.press/v119/zhang20ae.html} {{PEGASUS}: Pre-training with extracted gap-sentences for abstractive summarization}.
\newblock In \emph{Proceedings of the 37th International Conference on Machine Learning}, volume 119, pages 11328--11339.

\bibitem[{Zhang et~al.(2020{\natexlab{b}})Zhang, Sheng, Alhazmi, and Li}]{zhang2020adversarial}
Wei~Emma Zhang, Quan~Z. Sheng, Ahoud Alhazmi, and Chenliang Li. 2020{\natexlab{b}}.
\newblock \href {https://doi.org/10.1145/3374217} {Adversarial attacks on deep-learning models in natural language processing: A survey}.
\newblock \emph{ACM Transactions on Intelligent Systems and Technology}, 11(3).

\bibitem[{Zou et~al.(2025)Zou, Geng, Wang, and Jia}]{zou2025poisonedrag}
Wei Zou, Runpeng Geng, Binghui Wang, and Jinyuan Jia. 2025.
\newblock \href {https://www.usenix.org/conference/usenixsecurity25/presentation/zou-poisonedrag} {{PoisonedRAG}: Knowledge corruption attacks to retrieval-augmented generation of large language models}.
\newblock In \emph{34th USENIX Security Symposium (USENIX Security 25)}.

\end{thebibliography}

\appendix

\section{Methodology for Literature Compilation}
We detail the search and selection strategies used to curate the foundational content for this survey.

\subsection{Search Strategy} 
Initially, we conducted a comprehensive search in Google Scholar\footnote{https://scholar.google.com/}. We focused on high-recognized Natural Laugage Processing relevant venues such as ACL, EMNLP, NAACL, and TACL, and included the annual workshop on Fact Extraction and VERification (FEVER\footnote{https://fever.ai/index.html}) organized by the community specialized in fact-checking tasks. Besides, we took into consideration prestigious AI-related venues like AAAI; top-tier machine learning related venues like ICML, NeurIPS, and ICLR; and conferences in security such as USENIX Security and S\&P.

The search involved keywords including (1) fact-checking survey/review and fact-verification survey/review to compare current surveys with ours; (2) fact-checking attack and adversarial attack for works focusing on attack against fact-checking or fact-verification. 
Furthermore, for fact-checking defenses, we included publications in latest five years, which are highly cited and referenced as the state-of-the-art works. 

\subsection{Selection Strategy}
We only selected papers that directly target the subject matter of FC and attacks against FC. The selection was based on a careful review of the
abstract, introduction, conclusion, and limitations of each paper. Following the selection criteria, 50+ relevant papers on adversarial attack techniques, defenses, and evaluation (e.g., datasets and test FC models) were chosen to contribute the foundational content of this paper.

\section{From manual FC to automated FC}
\label{appx:manual2auto}
In practice, fact-checking websites such as Snopes\footnote{https://www.snopes.com/} and PolitiFact\footnote{https://www.politifact.com/}, employ human fact-checkers to verify claims and provide supporting evidence. In addition, Google Fact Check Tools\footnote{https://toolbox.google.com/factcheck/} serves as a resource that aggregates fact checks from reputable sources and provides ratings on the veracity of claims \cite{yang2024factchecking}.
These applied fact-checkers bring human expertise to fact-checking processes but encounter several critical challenges stemming from the nature of information, the adaptability of misinformation creators, and resource limitations:
1) \textit{Sheer volume of information}: The overwhelming scale and rapid spread of online content make it exceedingly difficult for human fact-checkers to keep up \cite{guo-etal-2022-survey}.
2) \textit{High complexity of misinformation}: Misinformation is often nuanced, interwoven with partial truths or framed in ways that make it challenging to debunk without substantial context or specialized expertise.
3) \textit{Multimodal content}: The increasing prevalence of multimedia, such as images, videos, and audio, complicates the verification process, as analyzing these multiple formats often requires specialized tools and skills \cite{akhtar-etal-2023-multimodal}; 
4) \textit{Evolving tactics}: Misinformation creators continually adapt their methods, including the use of AI-generated content \cite{kim-etal-2021-robust,abdelnabi2023fact,przybyła2024attacking}, making it harder for human fact-checkers to stay ahead.
5) \textit{Resource constraints}: Fact-checking organizations often operate with limited resources, including insufficient staff, inadequate funding, and limited access to relevant information, making it difficult to meet the demands for scalability, accuracy, and efficiency.

\section{Benchmark Datasets and Evaluation Criteria}
\label{appx:evaluation}

\subsection{Attack Target Dataset}
This section introduces 15 source datasets used to evaluate current adversarial attacks against AFC systems and their defenses. For a broader interest in datasets related to FC tasks, we refer readers to \cite{guo-etal-2022-survey}, \cite{akhtar-etal-2023-multimodal}, and \cite{eldifrawi-etal-2024-automated}.

\begin{itemize}
    \item The FEVER (a.k.a. FEVER 1.0) dataset \cite{thorne-etal-2018-fever} consists of 185,445 claims generated by altering sentences extracted from Wikipedia and subsequently verified without knowledge of the sentence they were derived from. 

    \item The FEVER-adv dataset is composed of 1,000 adversarial claims provided by adversaries in \cite{thorne-etal-2019-evaluating}.

    \item The FEVER 2.0 dataset \cite{thorne-etal-2019-fever2} consists of 1,174 claims created by the submissions of participants in the Breaker phase of the 2019 shared task. 
    It includes 1,000 adversarial claims provided by \citet{thorne-etal-2019-evaluating} and additional claims from participants. Only novel claims not contained in the original FEVER 1.0 dataset are included to form the dataset.

    \item The FEVER-sym dataset consists of synthetic claim-evidence pairs expended from the FEVER 1.0 dataset to challenge FC models trained on the dataset with claim-only bias \cite{schuster2019towards}. These synthetic pair holds the same relation (i.e. SUPPORTS or REFUTES) while expressing a fact that contradicts the original sentences. Combining the original and generated pairs, two new cross pairs that hold the inverse relations are obtained to be involved.
        
    \item The SciFact dataset \cite{wadden-etal-2020-fact} contains 1,409 expert-annotated scientific claims associated with 5,183 paper abstracts. 
    It presents the challenge of understanding scientific writing as systems must retrieve relevant sentences from paper abstracts and identify if the sentences support or refute a target scientific claim. It has emerged as a popular benchmark for evaluating scientific fact verification systems \cite{pradeep-etal-2021-scientific}.

    \item The CovidFact dataset \cite{saakyan-etal-2021-covid} contains totally 4,086 claims concerning the COVID-19 pandemic topic and associated evidence. Among them, 1,296 crowdsourced supported claims were crawled and filtered from the \textit{/r/COVID19} subreddit.
    The corresponding evidence is composed of documents originally provided with these claims when posted on the subreddit, along with resources retrieved through Google Search queries. Additionally, 2,790 refuted claims were automatically-generated by altering key words in the original supported claims.
    
    \item The CHEF dataset \cite{hu-etal-2022-chef} consists of 10,000 real-world Chinese claims, collected from 6 Chinese fact-checking websites covered multiple domains ranging from politics to public health. These claims are paired with annotated evidence retrieved from the Internet.

    \item The Factual-NLI dataset \cite{samarinas2020latent} comprises claim-evidence pairs from the FEVER dataset \cite{thorne-etal-2018-fever}, along with synthetic examples derived from the Natural Questions dataset \cite{kwiatkowski-etal-2019-natural} and the MS MARCO dataset \cite{nguyen2016msmarco} formatted as question-passage-answer triples. It initially has 911,146 claims for training and 86,543 for evaluation. Its noisy variant, the Factual-NLI+ dataset, introduces adversarial evidence: for every claim from FEVER, the top 30 web results are retrieved via Bing and evidence sentences with the highest BM25 score that are classified as neutral by the entailment model are retained. For claims generated from MS MARCO queries, irrelevant evidence retrieved from the MS MARCO dataset is included.

    \item The FoolMeTwice (FM2) dataset \cite{eisenschlos-etal-2021-fool} is a large dataset of adversarial entailment pairs collected through a fun multi-player game, containing fewer bigrams that ``give away'' the label on both the train and dev set. It has only REF and SUP claims manually written by human authors.

    \item The VitaminC dataset \cite{schuster-etal-2021-get} is a large-scale dataset based on factual revisions to Wikipedia. Following the setting of \cite{schuster2019towards}, two symmetric claims with opposing facts were generated for each revision when feasible, yielding a total of 325,724 claim-evidence pairs, and additional 163,180 pairs were generated following the synthetic process.

    \item The SufficientFacts dataset \cite{atanasova2022fact} is derived from subsets of the test sets of FEVER \cite{thorne-etal-2018-fever}, VitaminC \cite{schuster-etal-2021-get}, and HoVer \cite{jiang-etal-2020-hover}. Each instance includes the original claim and a modified version of the gold evidence with some information omitted. If the omitted information are deemed important by the majority of annotators, the label is changed to NEI; otherwise, the original label is retained. The dataset focuses on evaluating whether the remaining evidence is still sufficient for claim verification.

    \item The HoVer dataset \cite{jiang-etal-2020-hover} collects 26,171 multi-hop claims that require two to four evidence sentences  across diverse Wikipedia articles for verification. Each claim is manually annotated with a binary classification label: 15,023 are categorized as Supported, and 11,148 as Not-Supported.

    \item The DeSePtion dataset extends the FEVER 2.0 dataset \cite{thorne-etal-2019-fever2} by incorporating multiple adversarial attacks from Conjunction to Lexial Sub. in \cite{hidey-etal-2020-deseption}. 

    \item The PolitiHop-sym dataset \cite{zhang2024causal} combines the debiasing samples generated by GPT-4 and the original samples from the PolitiHop dataset \cite{ostrowski2021multi}, a small-scale multi-hop fact-checking dataset, including 500 manually annotated claims. Each PolitiHop claim is linked to a PolitiFact\footnote{https://www.politifact.com/} article containing a professional fact-checker's analysis and veracity assessment, from which sufficient sets of evidence sentences were selected.

    \item The AVeriTeC dataset \cite{schlichtkrull2023AVeriTeC} is a collection of 4,568 real-world claims fact-checked by 50 organizations. Each claim features question-and-answer pairs grounded in online evidence, along with textual justifications explaining how the evidence supports a verdict. Claims are categorized into one of four classes: Supported, Refuted, Not Enough Evidence, or Conflicting Evidence/Cherry-picking. 
    
\end{itemize}


\subsubsection{Insights into Evaluation}
\noindent
\textbf{Extending to Multimodal Data.} 
The current datasets involved in deploying or performing attacks against AFC systems are exclusively focused on single-modal data, specifically text data. Future efforts are expected to be made on multimodal data \cite{akhtar-etal-2023-multimodal}. 

\noindent
\textbf{Broadening Evaluation Domains and Realistic Settings.} The vulnerabilities of AFC systems are limited to being tested on data sourced from Wikipedia pages, scientific content, and social media. More domains should be involved in effectiveness evaluation of adversarial attacks. In parallel, since most evaluations rely on a few datasets like FEVER (see Tables~\ref{tab:adv_pair_eval}, \ref{tab:adv_claim_eval}, and \ref{tab:adv_evidence_eval} in Appendix~\ref{appx:summary}), there is a need to assess AFC robustness in more realistic settings. These include complex claims requiring decomposition \cite{hu-etal-2025-decomposition} and real-world claims paired with high-quality, web-retrieved evidence that ensures sufficiency and avoids temporal leakage \cite{schlichtkrull2023AVeriTeC}, where adversarial vulnerabilities may manifest differently.

\subsection{Evaluation Metrics}
\label{subsec:evaluation}

\subsubsection{Evaluation on Claim Verification}
\begin{itemize}

\item \textbf{FEVER Score} \cite{thorne-etal-2018-fact}: accuracy for claim verification, conditioned on retrieving at least one complete supporting evidence set for SUP and REF claims;

\item \textbf{Correctness Rate} \cite{thorne-etal-2019-evaluating}: percentage of grammatically correct and coherent claims with correct labels and evidence;

\item \textbf{Potency} \cite{thorne-etal-2019-evaluating}: average reduction (from 1) in FEVER score across FC systems;

\item \textbf{Adjusted Potency} \cite{thorne-etal-2019-evaluating}: weighted average reduction in FEVER score across FC systems, adjusted by correctness rate;

\item \textbf{Accuracy} \cite{thorne-etal-2018-fact}: percentage of claims with correctly predicted veracity labels;

\item \textbf{Resilience} \cite{thorne-etal-2019-evaluating}: average FEVER score scaled by correctness rate, where the weights correspond to the correctness rate for each adversary;

\item \textbf{Macro-F1} \cite{atanasova-etal-2020-generating}: unweighted average of the F1 scores computed for each claim class;

\item \textbf{`→ NEI' (\%)} \cite{abdelnabi2023fact}: ratio of verdict predictions that change to NEI;

\item \textbf{Attack Success Rate} \cite{he2025fact2fiction}: proportion of target claims where the attack successfully induces the intended verdict inversion;

\item \textbf{System Fail Rate} \cite{he2025fact2fiction}: proportion of target claims where the attack causes the fact-checking system to produce an incorrect verdict.

\subsubsection{Evaluation on Evidence Retrieval}
\textbf{Document Recall} \cite{kim-etal-2021-robust}: percentage of ground-truth documents (e.g., Wikipedia page in FEVER datasets) correctly retrieved by the fact-checking system;

\noindent
\textbf{Evidence Recall} \cite{thorne-etal-2019-evaluating}: percentage of ground-truth evidence sentences correctly retrieved by the fact-checking system;

\noindent
\textbf{Adversarial Evidence Recall} \cite{abdelnabi2023fact}: percentage of adversarial evidence sentences retrieved by the fact-checking system.

\noindent
\textbf{Successful Injection Rate} \cite{he2025fact2fiction}: proportion of malicious evidences among all retrieved evidences.

\end{itemize}

\subsubsection{Evaluation of Adversarial Claim Quality}

\begin{itemize}

\item \textbf{Perplexity} \cite{atanasova-etal-2020-generating}: perplexity of claims with prepended triggers measures how unpredictable a language model finds the modified claims compared to natural text;

\item \textbf{Semantic Similarity} \cite{atanasova-etal-2020-generating}: averaged semantic similarity between each claim and associated triggers;

\item \textbf{Claim quality} \cite{atanasova-etal-2020-generating}: overall quality of generated claims assessed by two independent annotators on a 1-5 scale.

\end{itemize}

\section{Figures for Adversarial Attack Overview}
\label{appx:figures}
\begin{figure}[htbp]
    \centering
    \includegraphics[width=0.81\linewidth]{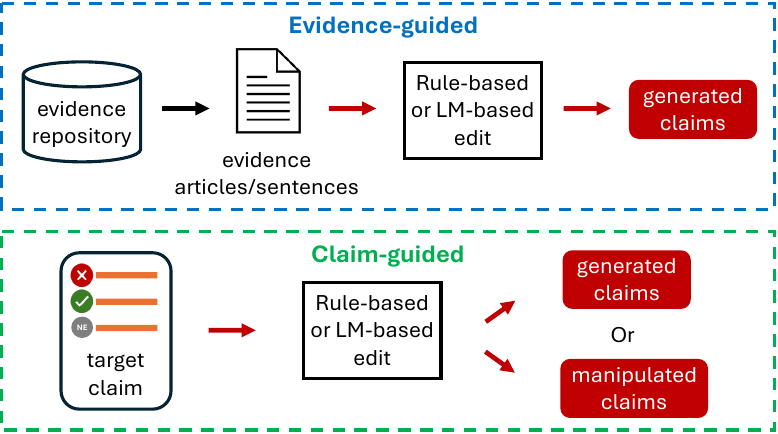}
    \caption{Overview of adversarial claim attacks: evidence-guided and claim-guided.}
    \label{fig:adv_claim}
\end{figure}

\begin{figure}[htbp]
    \centering
    \includegraphics[width=0.81\linewidth]{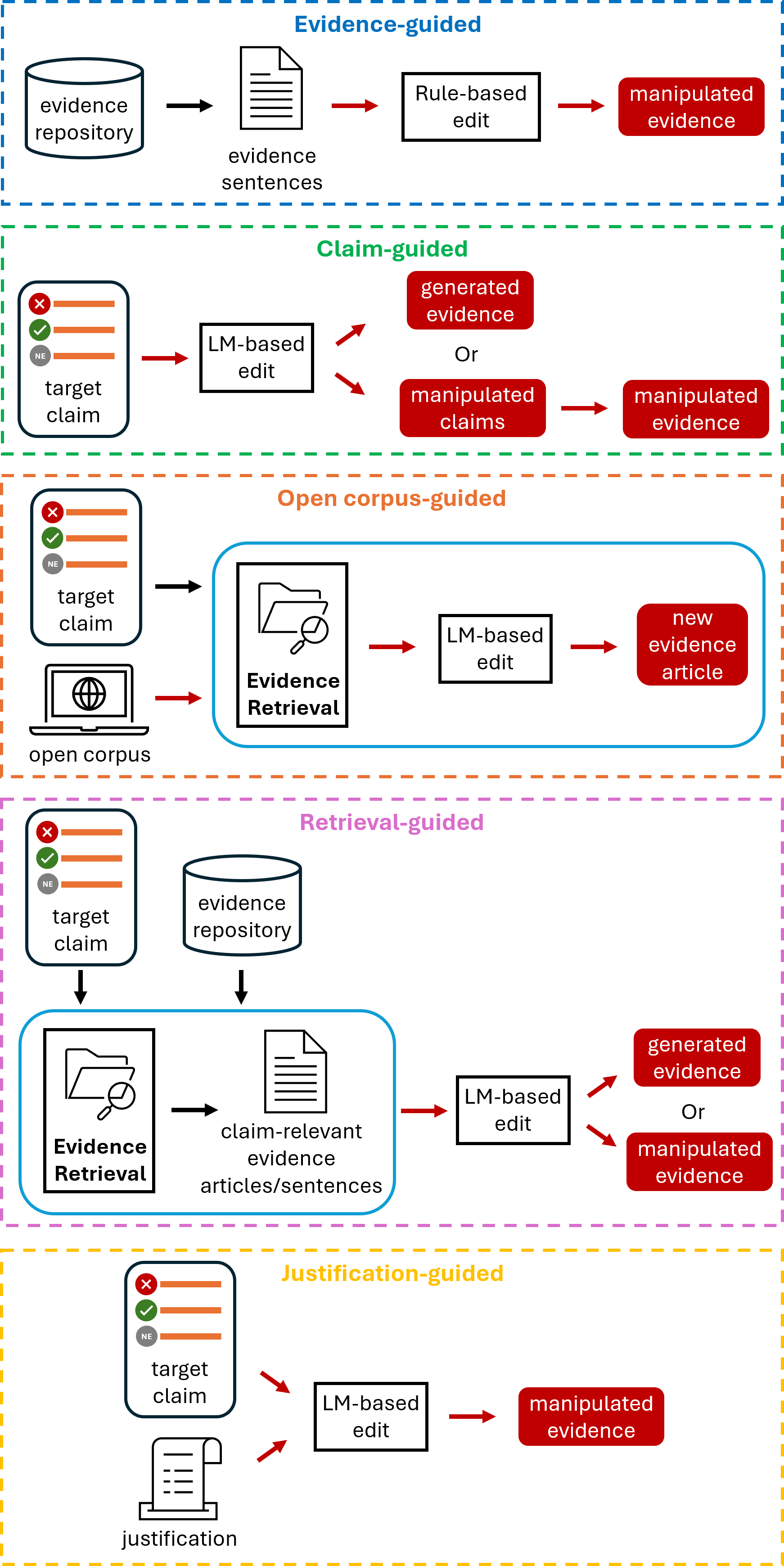}
    \caption{Overview of adversarial evidence attacks: evidence-guided, claim-guided, open corpus-guided, retrieval-guided, and justification-guided.}
    \label{fig:adv_evidence}
\end{figure}

\begin{figure}[htbp]
    \centering
    \includegraphics[width=1\linewidth]{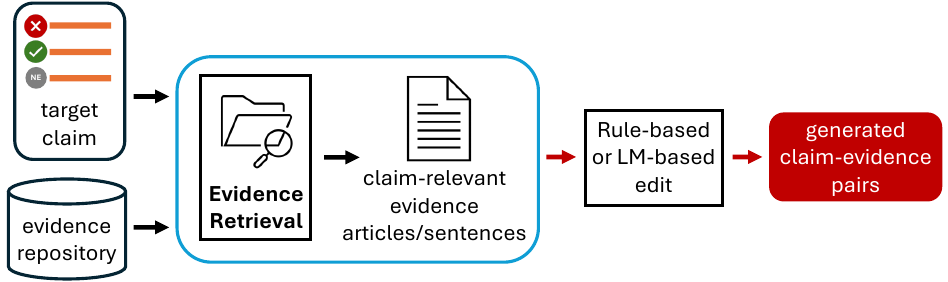}
    \caption{Overview of adversarial claim-evidence pair attacks.}
    \label{fig:adv_pair}
\end{figure}

We provide Figs.~\ref{fig:adv_claim} to \ref{fig:adv_pair} as complementary for our structured overview of current adversarial attacks against AFC systems in Sec.~\ref{sec:FC_attack}

\section{Characteristics of Adversarial Attacks}
\label{appx:tables}
To support a more comprehensive comparison of adversarial attacks from both security and technique perspectives, Tables~\ref{tab:adv_claim} to \ref{tab:adv_evidence_claim_pair} present key characteristics: direct attack target, method, source dataset (i.e., which dataset a certain attack were performed on), and access setting (black-box vs. white-box) for FC retrieval and verification models. The attack target notation (e.g., ``SUP → REF'') indicates a deliberate verdict shift from the gold label to a target label, while ``Generic'' signifies no specific target verdict.

\section{Summary Tables of Adversarial Attacks' Performance}
\label{appx:summary}
While ensuring a fair comparison across existing studies is challenging due to different settings, we provide summary tables of adversarial attacks, which offer an approximate view of performance differences. For clarity, we focus on evaluations conducted on FEVER 1.0 for both claim attacks (Table~\ref{tab:adv_claim_eval}) and evidence attacks (Table~\ref{tab:adv_evidence_eval}). Since the category of evidence–claim attacks comprises only two cases, we report both in 
Table~\ref{tab:adv_pair_eval}.

\begin{table}[htbp]
\centering
\begin{threeparttable}
\resizebox{0.5\textwidth}{!}{%
\begin{tabular}{ccccccc}
\toprule
\multirow{2}{*}{\textbf{Attack Technique}} & \multirow{2}{*}{\textbf{Target Dataset}} & \multirow{2}{*}{\textbf{Test FC System}} & \multicolumn{2}{c}{\textbf{Acc}} & \multicolumn{2}{c}{\textbf{Macro-F1}} \\  
\cmidrule(lr){4-5} \cmidrule(lr){6-7} 
&  &  & Org & Gen & Org & Gen \\ \midrule 
\multirow{3}{*}{Symmetric} & \multirow{3}{*}{FEVER 1.0} & NSMN & 81.8 & 58.7 & -- & -- \\
& & ESIM & 80.8 & 55.9 & -- & -- \\
& & GPT-3.5-Turbo & 86.2 & 58.3 & -- & -- \\
\midrule
\multirow{8}{*}{GPT-4 Symmetric} & \multirow{8}{*}{CHEF} & GPT-4 + DeBERTa & 77.34 & 43.68 & 75.26 & 44.57 \\
& & mDeBERTa & 82.45 & 53.27 & 80.68 & 51.09 \\
& & GPT-3.5-Turbo & 80.00 & 53.73 & 55.25 & 36.78 \\
& & BERT-base & 76.35 & 38.56 & 75.36 & 37.62 \\
& & Attention-based & 78.96 & 39.98 & 78.12 & 39.62 \\
& & KGAT & 79.55 & 39.61 & 76.97 & 38.67 \\
& & Chinese DeBERTa & 86.69 & 57.84 & 84.98 & 54.31 \\
& & GPT-4-Turbo & 85.60 & 65.20 & 60.70 & 47.12 \\
\bottomrule
\end{tabular}%
}                
\begin{tablenotes}
\tiny
\item \textbf{Acc}: Accuracy; \textbf{Org}: Original dataset pairs; \textbf{Gen}: Generated claim-evidence\\ pairs; \textbf{--}: This metric was not assessed.
\item \underline{Test FC systems}: NSMN~\cite{Nie_Chen_Bansal_2019}; ESIM~\cite{gardner-etal-2018-allennlp};\\ GPT-3.5-Turbo~\cite{openai2022gpt3-5}; GPT-4, GPT-4-Turbo~\cite{openai2024gpt4technicalreport}; DeBERTa\\~\cite{he2021deberta}; mDeBERTa~\cite{he2023debertav3}; BERT~\cite{devlin-etal-2019-bert};\\ Attention-based~\cite{gupta-srikumar-2021-x}; KGAT~\cite{liu-etal-2020-fine};\\ Chinese DeBERTa~\cite{zhang2023fengshenbang}.
\end{tablenotes}
\end{threeparttable}
\caption{Evaluation of adversarial claim-evidence pair attacks’ performance against test FC systems on FEVER 1.0 or CHEF.}
\label{tab:adv_pair_eval}
\end{table}

\begin{table*}[htbp]
\centering
\begin{threeparttable}
\resizebox{\textwidth}{!}{%
\begin{tabular}{cclccccc}
\toprule
\multirow{2}{*}{\textbf{Attack Technique}} & \multirow{2}{*}{\textbf{Target Dataset}} & \multicolumn{5}{c}{\textbf{Attack Settings}} & \multirow{2}{*}{\textbf{Attack Target}\tnote{\textcolor{red}{1}}} \\ \cmidrule{3-7}
&  & \multicolumn{1}{c}{\textbf{Attack Method}} & \textbf{BB-Ret} & \textbf{WB-Ret} & \textbf{BB-Ver} & \textbf{WB-Ver} & \\ \midrule
\multicolumn{8}{c}{\textbf{Rule-based Approaches}} \\ \midrule
\rowcolor[HTML]{C0C0C0} 
\begin{tabular}[c]{@{}c@{}}Model-targeting\\ \cite{thorne-etal-2019-evaluating}\end{tabular} & FEVER 1.0 & \begin{tabular}[c]{@{}l@{}}Generate meaning-preserving transformed claims guided by model predictions\\ using Semantically Equivalent Adversarial Rules (SEARs) and \cite{Nie_Chen_Bansal_2019} \end{tabular} & \ding{51} &  & \ding{51} & \ding{51}\tnote{\textcolor{cyan}{2}} & \begin{tabular}[c]{@{}c@{}}SUP → REF\\ REF → SUP\end{tabular} \\
\begin{tabular}[c]{@{}c@{}}Dataset Bias\\ \cite{thorne-etal-2019-evaluating}\end{tabular} & FEVER 1.0 & \begin{tabular}[c]{@{}l@{}}Introduce out-of-distribution patterns into adversarial claims via transformations: \\ entailment preserving rewrites,
simple negations, and complex negations\end{tabular} & \ding{51} &  & \ding{51} &  & \begin{tabular}[c]{@{}c@{}}SUP → REF\\ REF → SUP\end{tabular} \\
\rowcolor[HTML]{C0C0C0} 
\begin{tabular}[c]{@{}c@{}}EntityLess\\ \cite{kim-allan-2019-fever}\end{tabular} & FEVER 1.0 & Retain only generic entity terms in manipulated claims & \ding{51} &  & \ding{51} &  & \begin{tabular}[c]{@{}c@{}}Disrupted \\ retrieval \end{tabular} \\
\begin{tabular}[c]{@{}c@{}}EntityLinking\\ \cite{kim-allan-2019-fever}\end{tabular} & FEVER 1.0 & Replace entity names in claims with uncommon alternatives & \ding{51} &  & \ding{51} &  & \begin{tabular}[c]{@{}c@{}}Disrupted \\ retrieval \end{tabular} \\
\rowcolor[HTML]{C0C0C0} 
\begin{tabular}[c]{@{}c@{}}Controversy\\ \cite{kim-allan-2019-fever}\end{tabular} & FEVER 1.0 & Generate REF claims associated with conflicting evidence sentence pairs & \ding{51} &  & \ding{51} &  & REF → SUP \\
\begin{tabular}[c]{@{}c@{}}NotClear\\ \cite{kim-allan-2019-fever}\end{tabular} & FEVER 1.0 & Generate NEI claims based on evidence sentences with the phrase ``not clear'' & \ding{51} &  & \ding{51} &  & NEI → SUP/REF \\
\rowcolor[HTML]{C0C0C0} 
\begin{tabular}[c]{@{}c@{}}FiniteSet\\ \cite{kim-allan-2019-fever}\end{tabular} & FEVER 1.0 & Generate claims guided by temporal reasoning for time-dependent events & \ding{51} &  & \ding{51} &  & \begin{tabular}[c]{@{}c@{}}SUP → NEI\\ REF → NEI \end{tabular} \\
\begin{tabular}[c]{@{}c@{}}SubsetNum\\ \cite{kim-allan-2019-fever}\end{tabular} & FEVER 1.0 & Generate SUP claims guided by subset reasoning over entity relationships & \ding{51} &  & \ding{51} &  & SUP → NEI \\
\rowcolor[HTML]{C0C0C0} 
\begin{tabular}[c]{@{}c@{}}NotEnoughInfo Add\\ \cite{kim-allan-2019-fever}\end{tabular} & FEVER 1.0 & Add NEI claims to balance adversarial classes  & \ding{51} &  & \ding{51} &  & NEI → REF/SUP \\
\begin{tabular}[c]{@{}c@{}}Conjunction\\ \cite{hidey-etal-2020-deseption}\end{tabular} & FEVER 1.0 & Generate claims with clauses from single Wikipedia pages & \ding{51} &  & \ding{51} &  & \begin{tabular}[c]{@{}c@{}}REF → SUP/NEI\\ NEI → SUP/REF\end{tabular} \\
\rowcolor[HTML]{C0C0C0} 
\begin{tabular}[c]{@{}c@{}}Multi-hop\\ \cite{hidey-etal-2020-deseption}\end{tabular} & FEVER 1.0 & Generate claims with clauses across diverse Wikipedia pages  & \ding{51} &  & \ding{51} &  & \begin{tabular}[c]{@{}c@{}}Disrupted \\ retrieval \end{tabular} \\
\begin{tabular}[c]{@{}c@{}}Add. Unver.\\ \cite{hidey-etal-2020-deseption}\end{tabular} & FEVER 1.0 & Generate NEI claims with additional unverifiable propositions & \ding{51} &  & \ding{51} &  & NEI  → REF/SUP \\
\rowcolor[HTML]{C0C0C0} 
\begin{tabular}[c]{@{}c@{}}Date Manip.\\ \cite{hidey-etal-2020-deseption}\end{tabular} & FEVER 1.0 & \begin{tabular}[c]{@{}l@{}} Manipulate claims with date manipulation heuristics: arithmetic, range,\\ and verbalization  \end{tabular} & \ding{51} &  & \ding{51} &  & Generic \\
\begin{tabular}[c]{@{}c@{}}Multi-hop Temp.\\ \cite{hidey-etal-2020-deseption}\end{tabular} & FEVER 1.0 & Generate claims with temporal relations linking entities across Wikipedia pages  & \ding{51} &  & \ding{51} &  & \begin{tabular}[c]{@{}c@{}}Disrupted \\ retrieval \end{tabular} \\
\rowcolor[HTML]{C0C0C0} 
\begin{tabular}[c]{@{}c@{}}Entity Disamb.\\ \cite{hidey-etal-2020-deseption}\end{tabular} & FEVER 1.0 & Generate claims containing ambiguous entities & \ding{51} &  & \ding{51} &  & \begin{tabular}[c]{@{}c@{}}Disrupted \\ retrieval \end{tabular} \\
\begin{tabular}[c]{@{}c@{}}Lexical Sub.\\ \cite{hidey-etal-2020-deseption}\end{tabular} & FEVER 1.0 & \begin{tabular}[c]{@{}l@{}}Manipulate claims by applying genetic algorithm to replace synonyms, hypernyms, \\ or hyponyms \end{tabular} & \ding{51} &  & \ding{51} &  & Generic \\
\rowcolor[HTML]{C0C0C0} 
\begin{tabular}[c]{@{}c@{}}Game-play\\ \cite{eisenschlos-etal-2021-fool} \end{tabular} & FM2 & Handcraft plausible but hard claims through a multi-player game & \ding{51} &  & \ding{51} &  & \begin{tabular}[c]{@{}c@{}}SUP → REF/NEI\\
REF → SUP/NEI\end{tabular} \\ 
\begin{tabular}[c]{@{}c@{}}Contractions\\ \cite{factual2025mamta} \end{tabular} & FEVER 1.0 & Replace words in claims with contractions (e.g., \textit{do not} → \textit{don't})  & \ding{51} &  & \ding{51} &  & Generic \\ 
\rowcolor[HTML]{C0C0C0} 
\begin{tabular}[c]{@{}c@{}}Expansions\\ \cite{factual2025mamta} \end{tabular} & FEVER 1.0 & Replace a contraction in claims with its full form (e.g., \textit{don't} → \textit{do not}) & \ding{51} &  & \ding{51} &  & Generic \\ 
\begin{tabular}[c]{@{}c@{}}Jumbling\\ \cite{factual2025mamta} \end{tabular} & FEVER 1.0 & Perturb a claim by randomly changing the order of its words & \ding{51} &  & \ding{51} &  & Generic \\ 
\rowcolor[HTML]{C0C0C0} 
\begin{tabular}[c]{@{}c@{}}Num to Words\\ \cite{factual2025mamta} \end{tabular} & FEVER 1.0 & Convert all numerical digits in claims to their word equivalents (e.g., \textit{2} → \textit{two}) & \ding{51} &  & \ding{51} &  & Generic \\ 
\begin{tabular}[c]{@{}c@{}}Repeat Phrases\\ \cite{factual2025mamta} \end{tabular} & FEVER 1.0 & Append the first quarter of the claim to the end of the original claim & \ding{51} &  & \ding{51} &  & Generic \\
\rowcolor[HTML]{C0C0C0} 
\begin{tabular}[c]{@{}c@{}}Subject Verb Disag.\\ \cite{factual2025mamta} \end{tabular} & FEVER 1.0 & Introduce subject-verb agreement errors (i.e., singular vs plural) into claims & \ding{51} &  & \ding{51} &  & Generic \\ 
\begin{tabular}[c]{@{}c@{}}Typos\\ \cite{factual2025mamta} \end{tabular} & FEVER 1.0 & Manipulate claims by swapping adjacent characters & \ding{51} &  & \ding{51} &  & Generic \\ 
\rowcolor[HTML]{C0C0C0} 
\begin{tabular}[c]{@{}c@{}}Word Repetition\\ \cite{factual2025mamta} \end{tabular} & FEVER 1.0 & Duplicate a random selected word in the claim immediately after itself & \ding{51} &  & \ding{51} &  & Generic \\ 
\begin{tabular}[c]{@{}c@{}}Synonyms\\ \cite{factual2025mamta} \end{tabular} & FEVER 1.0 & Manipulate claims by replacing adjectives with synonyms from WordNet & \ding{51} &  & \ding{51} &  & Generic \\ 
\rowcolor[HTML]{C0C0C0} 
\begin{tabular}[c]{@{}c@{}}Tautology\\ \cite{factual2025mamta} \end{tabular} & FEVER 1.0 & Append \textit{and true is true} three times at the end of the claim & \ding{51} &  & \ding{51} &  & Generic \\ 
\begin{tabular}[c]{@{}c@{}}Phonetic Perturb.\\ \cite{factual2025mamta} \end{tabular} & FEVER 1.0 & \begin{tabular}[c]{@{}l@{}}Apply phonetic perturbations using a human-written dictionary with a word-level \\ perturbation budget \end{tabular} & \ding{51} &  & \ding{51} &  & Generic \\ 
\rowcolor[HTML]{C0C0C0} 
\begin{tabular}[c]{@{}c@{}}Character Swapping\\ \cite{factual2025mamta} \end{tabular} & FEVER 1.0 & Randomly swap adjacent characters within a word in the claim & \ding{51} &  & \ding{51} &  & Generic \\ 
\begin{tabular}[c]{@{}c@{}}Character Repetition\\ \cite{factual2025mamta} \end{tabular} & FEVER 1.0 & Randomly duplicate a non-initial/final character within a word in the claim & \ding{51} &  & \ding{51} &  & Generic \\ 
\rowcolor[HTML]{C0C0C0} 
\begin{tabular}[c]{@{}c@{}}Character Insertion\\ \cite{factual2025mamta} \end{tabular} & FEVER 1.0 & \begin{tabular}[c]{@{}l@{}}Randomly select a non-initial/final character from a word and insert it after the \\ selected character \end{tabular} & \ding{51} &  & \ding{51} &  & Generic \\ 
\begin{tabular}[c]{@{}c@{}}Character Deletion\\ \cite{factual2025mamta} \end{tabular} & FEVER 1.0 & Randomly delete a non-initial/final character within a word in the claim & \ding{51} &  & \ding{51} &  & Generic \\ 
\rowcolor[HTML]{C0C0C0} 
\begin{tabular}[c]{@{}c@{}}Homoglyph Perturb.\\ \cite{factual2025mamta} \end{tabular} & FEVER 1.0 & Replace characters with homoglyphs from the Unicode Security dictionary & \ding{51} &  & \ding{51} &  & Generic \\ 
\begin{tabular}[c]{@{}c@{}}LEET Perturb.\\ \cite{factual2025mamta} \end{tabular} & FEVER 1.0 & Replace letters with visually similar symbols using a predefined dictionary  & \ding{51} &  & \ding{51} &  & Generic \\ 
\midrule
\multicolumn{8}{c}{\textbf{LM-based Approaches}} \\ \midrule
\rowcolor[HTML]{C0C0C0} 
\begin{tabular}[c]{@{}c@{}}Lexically-informed\\ \cite{thorne-etal-2019-evaluating}\end{tabular} & FEVER 1.0 & \begin{tabular}[c]{@{}l@{}}Generate claims by replacing nouns and adjectives with WordNet synset \\ lemmas, followed by a neural back-translation paraphrasing model \end{tabular} & \ding{51} &  & \ding{51} &  & \begin{tabular}[c]{@{}c@{}}SUP → REF\\ REF → SUP\end{tabular} \\
\begin{tabular}[c]{@{}c@{}}Fact Mixing\\ \cite{niewinski-etal-2019-gem}\end{tabular} & FEVER 1.0 & \begin{tabular}[c]{@{}l@{}}Generate claims by controlling GPT-2 for text generation with novel target \\ vocabulary, mixing facts across articles\end{tabular} & \ding{51} &  & \ding{51} &  & Generic \\
\rowcolor[HTML]{C0C0C0} 
\begin{tabular}[c]{@{}c@{}}Adv. Trigger\\ \cite{atanasova-etal-2020-generating}\end{tabular} & FEVER 1.0 & \begin{tabular}[c]{@{}l@{}}Generate claims using GPT-2 fine-tuned with triggers optimized by extending \\ HotFlip with RoBERTa fine-tuned based on Semantic Textual Similarity\end{tabular} 
& \ding{51} &  & \ding{51} &  & Generic \\
\begin{tabular}[c]{@{}c@{}}Colloquial\\ \cite{kim-etal-2021-robust}\end{tabular} & FEVER 1.0 & Generate colloquial claims by BART-large fine-tuned on Wizard of Wikipedia  & \ding{51} &  & \ding{51} &  & \begin{tabular}[c]{@{}c@{}}Disrupted \\ retrieval \end{tabular} \\ \bottomrule
\end{tabular}%
}
\begin{tablenotes}
\tiny
\item \textbf{BB-Ret}: Black-box attack without access to the FC retrieval model's architecture; \textbf{WB-Ver}: White-box attack with access to the FC retrieval model's architecture; \\ 
\textbf{BB-Ver}: Black-box attack without access to the FC verification model's architecture; \textbf{WB-Ver}: White-box attack with access to the FC verification model's architecture; \\ 
``\ding{51}'' signifies that the corresponding access setting has been explored for the attack; a blank indicates it has not.
\textbf{NEI}: NotEnoughInfo; \textbf{SUP}: Supported; \textbf{REF}: Refuted. 
\item[\textcolor{red}{1}] Regarding the target attack, the notation ``A → B'' denotes a corrupted verdict in which the gold label A is intentionally altered to the target label B, and ``Generic'' means \\ no specified verdict shift for corrupted verdict.
\item [\textcolor{cyan}{2}] We indicate white-box verification access for NSMN, one of the test FC models, since the Model-targeting attack leverages its internal predictions to generate adversarial claims.
\end{tablenotes}
\end{threeparttable}
\caption{A summary of adversarial claim attack techniques, categorized into rule-based and LM-based approaches.}
\label{tab:adv_claim}
\end{table*}

\begin{table*}[htbp]
\centering
\begin{threeparttable}
\resizebox{\textwidth}{!}{%
\begin{tabular}{cclccccc}
\toprule
\multirow{2}{*}{\textbf{Attack Technique}} & \multirow{2}{*}{\textbf{Target Dataset}} & \multicolumn{5}{c}{\textbf{Attack Settings}} & \multirow{2}{*}{\textbf{Attack Target}\tnote{\textcolor{red}{1}}} \\ \cmidrule{3-7}
&  & \multicolumn{1}{c}{\textbf{Attack Method}} & \textbf{BB-Ret} & \textbf{WB-Ret} & \textbf{BB-Ver} & \textbf{WB-Ver} & \\ \midrule
\rowcolor[HTML]{C0C0C0} 
\begin{tabular}[c]{@{}c@{}}AdvAdd\tnote{\textcolor{cyan}{2}} \\ \cite{du_2022_synthetic}\end{tabular} & \begin{tabular}[c]{@{}c@{}}FEVER 1.0\\ SciFact\\ CovidFact\end{tabular} & Add adversarial evidence generated by Grover & \ding{51} &  & \ding{51} &  & \begin{tabular}[c]{@{}c@{}}Disrupted \\ retrieval\end{tabular} \\
\begin{tabular}[c]{@{}c@{}}AdvMod\\ \cite{du_2022_synthetic}\end{tabular} & FEVER 1.0 & Create adversarial evidence by PEGASUS for real-time verification & \ding{51} &  & \ding{51} & & \begin{tabular}[c]{@{}c@{}}Disrupted \\ retrieval\end{tabular} \\
\rowcolor[HTML]{C0C0C0} 
\begin{tabular}[c]{@{}c@{}}Imperceptible\\ \cite{boucher2022bad}\end{tabular} & FEVER 1.0 & \begin{tabular}[c]{@{}l@{}}Manipulate evidence characters by replacing with homoglyphs,\\ reordering, or deleting them;\\ Minimize correct claim label probability of BERT classifier\end{tabular} & \ding{51} & \ding{51} & \ding{51} & & \begin{tabular}[c]{@{}c@{}} SUP → REF\\ REF → SUP\end{tabular} \\
\begin{tabular}[c]{@{}c@{}}Imperceptible\textsubscript{Ret}\\ \cite{boucher2022bad}\end{tabular} & FEVER 1.0 & \begin{tabular}[c]{@{}l@{}}Manipulate evidence characters;\\ Minimize claim-specific evidence ranking score of retrieval model\end{tabular} & \ding{51} &  & \ding{51} & & \begin{tabular}[c]{@{}c@{}}Disrupted \\ retrieval\end{tabular} \\
\rowcolor[HTML]{C0C0C0} 
\begin{tabular}[c]{@{}c@{}}Lexical Variation\\ \cite{alzantot-etal-2018-generating}\end{tabular} & FEVER 1.0 & Create adversarial claim-paired evidence by RoBERTaBASE & \ding{51} &  & \ding{51} & & \begin{tabular}[c]{@{}c@{}} SUP → REF\\ REF → SUP\end{tabular} \\
\begin{tabular}[c]{@{}c@{}}Contextualized Replace\\ \cite{li-etal-2020-bert-attack}\end{tabular} & FEVER 1.0 & \begin{tabular}[c]{@{}l@{}}Pre-train BERT to replace masked salient words for\\ maximum classification drop\end{tabular} & \ding{51} & \ding{51} & \ding{51} & & \begin{tabular}[c]{@{}c@{}}SUP → REF\\ REF → SUP\end{tabular} \\
\rowcolor[HTML]{C0C0C0} 
\begin{tabular}[c]{@{}c@{}}Omitting Paraphrase\\ \cite{abdelnabi2023fact}\end{tabular} & FEVER 1.0 & \begin{tabular}[c]{@{}l@{}}Paraphrase evidence by PEGASUS;\\ Minimize claim-specific retrieval ranking\end{tabular} & \ding{51} & \ding{51} & \ding{51} & & \begin{tabular}[c]{@{}c@{}}Disrupted \\ retrieval\end{tabular} \\
\begin{tabular}[c]{@{}c@{}}Omitting Generate\\  \cite{abdelnabi2023fact}\end{tabular} & FEVER 1.0 & Fine-tune GPT-2 to create alternative evidence omitting key parts & \ding{51} &  & \ding{51} & & \begin{tabular}[c]{@{}c@{}}Disrupted \\ retrieval\end{tabular} \\
\rowcolor[HTML]{C0C0C0} 
\begin{tabular}[c]{@{}c@{}}Claim-aligned\\ Re-writing\\ \cite{abdelnabi2023fact}\end{tabular} & FEVER 1.0 & \begin{tabular}[c]{@{}l@{}}Mask top important tokens by BERT verification model;\\ T5 reconstructs masked supporting evidence; \\ Select Top-K maximizing SUP probability of BERT\end{tabular} & \ding{51} &  & \ding{51} & & REF → SUP \\
\begin{tabular}[c]{@{}c@{}}Claim-aligned\\ Re-writing\textsubscript{ret}\\ \cite{abdelnabi2023fact}\end{tabular} & FEVER 1.0 & \begin{tabular}[c]{@{}l@{}}Mask tokens for lowest retrieval scores;\\ T5 reconstructs masked supporting evidence;\\ Select evidence maximizing retrieval scores\end{tabular} & \ding{51} &  & \ding{51} & & \begin{tabular}[c]{@{}c@{}}Disrupted \\ retrieval\end{tabular} \\
\rowcolor[HTML]{C0C0C0} 
\begin{tabular}[c]{@{}c@{}}Supporting Generation\\ \cite{abdelnabi2023fact}\end{tabular} & FEVER 1.0 & \begin{tabular}[c]{@{}l@{}}Fine-tune GPT-2 to generate supporting evidence;\\ Select evidence maximizing SUP probability of BERT\end{tabular} & \ding{51} &  & \ding{51} & & \begin{tabular}[c]{@{}c@{}} REF → SUP\\ NEI → SUP\end{tabular} \\
\begin{tabular}[c]{@{}c@{}} Neutral Noise\\ \cite{samarinas2021improving}\end{tabular} & \begin{tabular}[c]{@{}c@{}}FEVER 1.0 \\ Factual-NLI\end{tabular} & Add top-scoring entailment-neutral documents retrieved by BM25 & \ding{51} &  & \ding{51} & & \begin{tabular}[c]{@{}c@{}}Disrupted \\ retrieval\end{tabular} \\ 
\rowcolor[HTML]{C0C0C0}
\begin{tabular}[c]{@{}c@{}}Omission Generation \\ \cite{atanasova2022fact}\end{tabular} & \begin{tabular}[c]{@{}c@{}}FEVER 1.0 \\ VitaminC\\ HoVer\end{tabular} & \begin{tabular}[c]{@{}l@{}}Remove sentence constructs from evidence text and preserve evidence's\\ stance towards the claim\end{tabular} & \ding{51} &  & \ding{51} & & \begin{tabular}[c]{@{}c@{}}Disrupted \\ retrieval\end{tabular} \\
\begin{tabular}[c]{@{}c@{}}Fact2Fiction \\ \cite{he2025fact2fiction}\end{tabular} & AVeriTeC & \begin{tabular}[c]{@{}l@{}}A Planner generates a targeted adversarial answer for each sub-question; \\ An Executor crafts tailored malicious evidence corpora to compromise\\ each sub-question.\end{tabular} & \ding{51} &  & \ding{51} & & \begin{tabular}[c]{@{}c@{}} SUP → REF\\ REF → SUP\end{tabular} \\
\bottomrule
\end{tabular}%
}                
\begin{tablenotes}
\tiny
\item \textbf{BB-Ret}: Black-box attack without access to the FC retrieval model's architecture; \textbf{WB-Ver}: White-box attack with access to the FC retrieval model's architecture; \\ 
\textbf{BB-Ver}: Black-box attack without access to the FC verification model's architecture; \textbf{WB-Ver}: White-box attack with access to the FC verification model's architecture; \\ 
``\ding{51}'' signifies that the corresponding access setting has been explored for the attack; a blank indicates it has not. \textbf{NEI}: NotEnoughInfo; \textbf{SUP}: Supported; \textbf{REF}: Refuted.
\item[\textcolor{red}{1}] Regarding the target attack, the notation ``A → B'' denotes a corrupted verdict in which the gold label A is intentionally altered to the target label B, and ``Generic'' means \\ no specified verdict shift for corrupted verdict.
\item[\textcolor{cyan}{2}] The AdvAdd attack is implemented as the Claim-conditioned Article Generation attack in \cite{abdelnabi2023fact}, disrupting the retrieval process.
\end{tablenotes}
\end{threeparttable}
\caption{A summary of adversarial evidence attack techniques.}
\label{tab:adv_evidence}
\end{table*}

\begin{table*}[htbp]
\centering
\begin{threeparttable}
\resizebox{\textwidth}{!}{%
\begin{tabular}{cclccccc}
\toprule
\multirow{2}{*}{\textbf{Attack Technique}} & \multirow{2}{*}{\textbf{Target Dataset}} & \multicolumn{5}{c}{\textbf{Attack Settings}} & \multirow{2}{*}{\textbf{Attack Target}\tnote{\textcolor{red}{1}}} \\ \cmidrule{3-7}
&  & \multicolumn{1}{c}{\textbf{Attack Method}} & \textbf{BB-Ret} & \textbf{WB-Ret} & \textbf{BB-Ver} & \textbf{WB-Ver} & \\ \midrule
\multicolumn{8}{c}{\textbf{Rule-based Approaches}} \\ \midrule
\rowcolor[HTML]{C0C0C0} 
\begin{tabular}[c]{@{}c@{}}Symmetric\\ \cite{schuster2019towards}\end{tabular} & FEVER 1.0 & \begin{tabular}[c]{@{}l@{}}Manually generate pairs holding the same relation but expressing a\\ different, contrary, fact\end{tabular} & \ding{51} &  & \ding{51} &  & Generic \\
\midrule
\multicolumn{8}{c}{\textbf{LM-based Approaches}} \\ \midrule
\rowcolor[HTML]{C0C0C0} 
\begin{tabular}[c]{@{}c@{}}GPT-4 Symmetric\\ \cite{zhang2024need}\end{tabular} & CHEF & \begin{tabular}[c]{@{}l@{}}Use GPT-4 to generate Chinese pairs holding the same relation but expressing\\ a different, contrary, fact\end{tabular}  & \ding{51} &  & \ding{51} &  & Generic \\
\bottomrule
\end{tabular}%
}
\begin{tablenotes}
\tiny
\item \textbf{BB-Ret}: Black-box attack without access to the FC retrieval model's architecture; \textbf{WB-Ver}: White-box attack with access to the FC retrieval model's architecture; \\ 
\textbf{BB-Ver}: Black-box attack without access to the FC verification model's architecture; \textbf{WB-Ver}: White-box attack with access to the FC verification model's architecture; \\ 
``\ding{51}'' signifies that the corresponding access setting has been explored for the attack; a blank indicates it has not.
\item[\textcolor{red}{1}] ``Generic'' means  no specified verdict shift for corrupted verdict.
\end{tablenotes}
\end{threeparttable}
\caption{A summary of adversarial claim-evidence pair attack techniques, categorized into rule-based and LM-based approaches.}
\label{tab:adv_evidence_claim_pair}
\end{table*}

\begin{table*}[ht!]
\centering
\begin{threeparttable}
\resizebox{\textwidth}{!}{%
\begin{tabular}{cccccccccccc|ccc}
\toprule
\multirow{2}{*}{\textbf{Attack Technique}}  & \multirow{2}{*}{\textbf{Test FC System}} & \multicolumn{2}{c}{\textbf{FEVER}} & \multicolumn{2}{c}{\textbf{Acc}} & \multirow{2}{*}{\textbf{Res}} & \multirow{2}{*}{\textbf{Macro-F1}} & \multicolumn{2}{c}{\textbf{Rec\textsubscript{Evd}}} & \multicolumn{2}{c|}{\textbf{Rec\textsubscript{Doc}}} & \multirow{2}{*}{\textbf{Corr}\tnote{\textcolor{cyan}{2}}} & \multirow{2}{*}{\textbf{Pot}\tnote{\textcolor{cyan}{2}}} & \multirow{2}{*}{\textbf{Adj}\tnote{\textcolor{cyan}{2}}} \\  
\cmidrule(lr){3-4} \cmidrule(lr){5-6} \cmidrule(lr){9-10} \cmidrule(lr){11-12} 
&  &  Org & Mod & Org & Mod &  &  & Org & Mod & Org & Mod &  &  & \\ \midrule 
\multirow{6}{*}{Model-targeting} & Papelo & 73.87 & 58.26 & -- & -- & 58.66\tnote{\textcolor{red}{1}} & -- & -- & -- & -- & -- & \multirow{6}{*}{62.5} & \multirow{6}{*}{57.84} & \multirow{6}{*}{36.15} \\
& NSMN & 68.77 & 47.85 & -- & -- & 51.09\tnote{\textcolor{red}{1}} & -- & -- & -- & -- & -- &  &  & \\
& HexaF & 67.37 & 50.15 & -- & -- & 50.06\tnote{\textcolor{red}{1}} & -- & -- & -- & -- & -- &  &  & \\
& Enhanced ESIM & 62.76 & 46.75 & -- & -- & 43.98\tnote{\textcolor{red}{1}} & -- & -- & -- & -- & -- &  &  & \\
& TF-IDF + ESIM & 32.83 & 28.13 & -- & -- & 26.86\tnote{\textcolor{red}{1}} & -- & -- & -- & -- & -- &  &  & \\
& TF-IDF + DA & 27.73 & 21.82 & -- & -- & 22.28\tnote{\textcolor{red}{1}} & -- & -- & -- & -- & -- &  &  & \\
\midrule
\multirow{8}{*}{Dataset Bias} & Oracle + DA & -- & -- & 82.38 & 62.56 & -- & -- & -- & -- & -- & -- & \multirow{8}{*}{89.5} & \multirow{8}{*}{63.16} & \multirow{8}{*}{56.53} \\ 
& Oracle + ESIM & -- & -- & 83.58 & 60.66 & -- & -- & -- & -- & -- & -- &  &  & \\
& Papelo & 73.87 & 56.36 & 75.58 & 57.26 & 58.66\tnote{\textcolor{red}{1}} & -- & 71.47 & 50.00 & -- & -- &  &  & \\
& NSMN & 68.77 & 48.85 & 70.27 & 50.35 & 51.09\tnote{\textcolor{red}{1}} & -- & 78.07 & 71.62 & -- & -- &  &  & \\
& HexaF & 67.37 & 45.25 & 74.67 & 51.35 & 50.06\tnote{\textcolor{red}{1}} & -- & 80.78 & 75.98 & -- & -- &  &  &  \\
& Enhanced ESIM & 62.76 & 31.53 & 67.56 & 36.84 & 43.98\tnote{\textcolor{red}{1}} & -- & 86.04 & 80.93 & -- & -- &  &  &  \\
& TF-IDF + ESIM & 32.83 & 20.72 & 49.75 & 37.34 & 26.86\tnote{\textcolor{red}{1}} & -- & 45.50 & 39.20 & -- & -- &  &  &  \\
& TF-IDF + DA & 27.73 & 18.32 & 46.49 & 37.44 & 22.28\tnote{\textcolor{red}{1}} & -- & 45.50 & 39.20 & -- & -- &  &  &  \\
\midrule
\multirow{6}{*}{Lexically-informed} & Papelo & 73.87 & 44.24 & -- & -- & 58.66\tnote{\textcolor{red}{1}} & -- & -- & -- & -- & -- & \multirow{6}{*}{34.0} & \multirow{6}{*}{65.64} & \multirow{6}{*}{22.32} \\
& NSMN & 68.77 & 39.44 & -- & -- & 51.09\tnote{\textcolor{red}{1}} & -- & -- & -- & -- & -- &  &  & \\
& HexaF & 67.37 & 35.64 & -- & -- & 50.06\tnote{\textcolor{red}{1}} & -- & -- & -- & -- & -- &  &  & \\
& Enhanced ESIM & 62.76 & 38.24 & -- & -- & 43.98\tnote{\textcolor{red}{1}} & -- & -- & -- & -- & -- &  &  & \\
& TF-IDF + ESIM & 32.83 & 27.83 & -- & -- & 26.86\tnote{\textcolor{red}{1}} & -- & -- & -- & -- & -- &  &  & \\
& TF-IDF + DA & 27.73 & 20.72 & -- & -- & 22.28\tnote{\textcolor{red}{1}} & -- & -- & -- & -- & -- &  &  & \\
\midrule
EntityLess & \multirow{7}{*}{\begin{tabular}[c]{@{}c@{}}FEVER 2.0 Shared\\ Task  Systems\end{tabular}} & -- & -- & -- & -- & -- & -- & -- & -- & -- & -- & \multirow{7}{*}{64.71} & \multirow{7}{*}{79.66} & \multirow{7}{*}{51.54} \\
EntityLinking &  & -- & -- & -- & -- & -- & -- & -- & -- & -- & -- &  &  & \\
Controversy &  & -- & -- & -- & -- & -- & -- & -- & -- & -- & -- &  &  & \\
NotClear &  & -- & -- & -- & -- & -- & -- & -- & -- & -- & -- &  &  & \\
FiniteSet &  & -- & -- & -- & -- & -- & -- & -- & -- & -- & -- &  &  & \\
SubsetNum &  & -- & 0.00\tnote{\textcolor{olive}{3}} & -- & 16.12\tnote{\textcolor{olive}{3}} & -- & -- & -- & -- & -- & -- &  &  & \\
NotEnoughInfo Add &  & -- & 76.39\tnote{\textcolor{olive}{3}} & -- & 76.39\tnote{\textcolor{olive}{3}} & -- & -- & -- & -- & -- & -- &  &  & \\
\midrule
Conjunction & \multirow{7}{*}{\begin{tabular}[c]{@{}c@{}}FEVER 2.0 Shared\\ Task  Systems\end{tabular}} & -- & 38.25\tnote{\textcolor{olive}{3}} & -- & 42.50\tnote{\textcolor{olive}{3}} & -- & -- & -- & -- & -- & -- & \multirow{7}{*}{81.44} & \multirow{7}{*}{68.51} & \multirow{7}{*}{55.79} \\
Multi-hop &  & -- & 31.54\tnote{\textcolor{olive}{3}} & -- & 51.64\tnote{\textcolor{olive}{3}} & -- & -- & -- & -- & -- & -- &  &  & \\
Add Unver &  & -- & 55.63\tnote{\textcolor{olive}{3}} & -- & 55.63\tnote{\textcolor{olive}{3}} & -- & -- & -- & -- & -- & -- &  &  & \\
Date Manip &  & -- & 27.53\tnote{\textcolor{olive}{3}} & -- & 34.18\tnote{\textcolor{olive}{3}} & -- & -- & -- & -- & -- & -- &  &  & \\
Multi-hop Temp &  & -- & 8.33\tnote{\textcolor{olive}{3}} & -- & 24.48\tnote{\textcolor{olive}{3}} & -- & -- & -- & -- & -- & -- &  &  & \\
Entity Disamb &  & -- & -- & -- & -- & -- & -- & -- & -- & -- & -- &  &  & \\
Lexical Sub &  & -- & 28.87\tnote{\textcolor{olive}{3}} & -- & 29.08\tnote{\textcolor{olive}{3}} & -- & -- & -- & -- & -- & -- &  &  & \\
\midrule
Fact Mixing & \begin{tabular}[c]{@{}c@{}}FEVER 2.0 Shared \\ Task Systems\end{tabular} & -- & 38.07\tnote{\textcolor{olive}{3}} & -- & 40.63\tnote{\textcolor{olive}{3}} & -- & -- & -- & -- & -- & -- & 84.81 & 78.80 & 66.83 \\
\midrule
Adv Trigger (w STS) & \multirow{2}{*}{RoBERTa} & -- & -- & -- & -- & -- & 63.5\tnote{\textcolor{green}{4}} & -- & -- & -- & -- & \multirow{2}{*}{--} & \multirow{2}{*}{--} & \multirow{2}{*}{--} \\
Adv Trigger (w/o STS)&  & -- & -- & -- & -- & -- &  53.4\tnote{\textcolor{green}{4}} & -- & -- & -- & -- &  &  & \\
\midrule
\multirow{6}{*}{Colloquial} & \scalebox{0.7}{KGAT (BERT) + Evidence Oracle} & -- & -- & 69.7 & 57.3 & -- & -- & -- & -- & -- & -- & \multirow{6}{*}{--} & \multirow{6}{*}{--} & \multirow{6}{*}{--} \\
& \scalebox{0.7}{KGAT (BERT) + WikiAPI + BERT} & 62.4 & 43.6 & 67.5 & 53.2 & -- & -- & 85.3 & 73.4 & 90.0 & 72.2 &  &  & \\
& \scalebox{0.7}{KGAT (BERT) + DPR + BERT} & 55.4 & 41.5 & 62.9 & 51.2 & -- & -- & 81.8 & 77.4 & 84.0 & 79.6 &  &  & \\
& \scalebox{0.7}{KGAT (CorefBERT) + Evidence Oracle} & -- & -- & 77.5 & 67.7 & -- & -- & -- & -- & -- & -- &  &  & \\
& \scalebox{0.7}{KGAT (CorefBERT) + WikiAPI + BERT} & 69.5 & 52.4 & 73.8 & 60.9 & -- & -- & 85.3 & 73.4 & 90.0 & 72.2 &  &  & \\
& \scalebox{0.7}{KGAT (CorefBERT) + DPR + BERT} & 52.4 & 55.4 & 61.1 & 61.0 & -- & -- & 81.8 & 77.4 & 84.0 & 79.6 &  &  & \\
\bottomrule
\end{tabular}%
}
\begin{tablenotes}
\tiny
\item \textbf{Acc}: Accuracy; \textbf{Res}: Resilience;  
\textbf{Rec\textsubscript{Evd}}: Evidence Sentence Recall; \textbf{Rec\textsubscript{Doc}}: Document Recall; \textbf{Pot:} Potency; \textbf{Corr:} Correctness Rate; \textbf{Adj}: Adjusted Potency; \textbf{Org}: Results on\\ original target claims; \textbf{Mod}: Results on modified claims; \textbf{STS}: Semantic Textual Similarity; \textbf{DPR}: Dense Passage Retrieval; \textbf{--}: This metric was not assessed.
\item \underline{Test FC systems}: Papelo~\cite{malon-2018-team}; NSMN~\cite{Nie_Chen_Bansal_2019}; HexaF~\cite{yoneda-etal-2018-ucl}; Enhanced ESIM, WikiAPI~\cite{hanselowski-etal-2018-ukp}; TF-IDF~\cite{chen-etal-2017-reading};\\ ESIM~\cite{gardner-etal-2018-allennlp}; DA~\cite{parikh-etal-2016-decomposable}; FEVER 2.0 Shared Task Systems~\cite{thorne-etal-2019-fever2}; RoBERTa~\cite{liu2019roberta};\\ KGAT~\cite{liu-etal-2020-fine}; BERT~\cite{devlin-etal-2019-bert}, CorefBERT~\cite{ye-etal-2020-coreferential}; DPR~\cite{karpukhin-etal-2020-dense}.
\item[\textcolor{red}{1}] Statistics for Resilience are with a mixture of adversarial claims generated by Model-targeting, Dataset Bias, and Lexically-informed attacks across test FC systems.
\item[\textcolor{cyan}{2}] In \cite{kim-allan-2019-fever}, \cite{hidey-diab-2018-team}, and \cite{niewinski-etal-2019-gem}, statistics for Corr, Pot, and Adj are with their respective attacks combined.
\item[\textcolor{olive}{3}] Statistics for average FEVER and Acc are reported for adversarial claims annotated as ``correct'' and attack types with over five generated claims, referring to \cite{thorne-etal-2019-fever2}.
\item[\textcolor{green}{4}] Statistics for Macro-F1 score are with verification results over generated claims.
\end{tablenotes}
\end{threeparttable}
\caption{Evaluation of adversarial claim attacks' performance against test FC systems on FEVER 1.0.}
\label{tab:adv_claim_eval}
\end{table*}

\begin{table*}[ht!]
\centering
\begin{threeparttable}
\resizebox{\textwidth}{!}{%
\begin{tabular}{ccccccccccccccc}
\toprule
\multirow{2}{*}{\textbf{Attack Technique}}  & \multirow{2}{*}{\textbf{Test FC System}} & \multicolumn{2}{c}{\textbf{Acc on All}} & \multicolumn{2}{c}{\textbf{Acc on SUP}}  & \multicolumn{2}{c}{\textbf{Acc on REF}} & \multicolumn{2}{c}{\textbf{Acc on NEI}} & \multicolumn{2}{c}{\textbf{Macro-F1}} & \multirow{2}{*}{\textbf{Rec$_{\text{Evd}}^{\text{Adv}}$}} & \multirow{2}{*}{\textbf{Rec$_{\text{Evd}}^{\text{Gold}}$}} & \multirow{2}{*}{\textbf{→ NEI}} \\  
\cmidrule(lr){3-4} \cmidrule(lr){5-6} \cmidrule(lr){7-8} \cmidrule(lr){9-10} \cmidrule(lr){11-12} 
&  &  Org & Mod & Org & Mod & Org & Mod & Org & Mod & Org & Mod &  & \\ \midrule 
\multirow{3}{*}{AdvAdd} 
& KGAT & 70.76 & 29.02 & -- & -- & 72.50 & 42.45 & 69.01 & 15.59 & -- & -- & -- & -- & -- \\
& CorefBERT & 73.05 & 28.59 & -- & -- & 74.03 & 39.63 & 72.07 & 17.54 & -- & -- & -- & -- & -- \\
& MLA & 75.92 & 51.86 & -- & -- & 78.71 & 71.84 & 73.13 & 31.87 & -- & -- & -- & -- & -- \\
\midrule
\multirow{3}{*}{AdvMod} 
& KGAT & 70.76 & 37.22 & -- & -- & 72.50 & 51.74 & 69.01 & 22.70 & -- & -- & -- & -- & -- \\
& CorefBERT & 73.05 & 32.62 & -- & -- & 74.03 & 36.66 & 72.07 & 28.58 & -- & -- & -- & -- & -- \\
& MLA & 75.92 & 52.72 & -- & -- & 78.71 & 70.57 & 73.13 & 34.86 & -- & -- & -- & -- & -- \\
\midrule
Imperceptible & & & & & & & & & & & & & & \\
\cmidrule(lr){1-2}
\multirow{4}{*}{Homoglyph ($\epsilon = 5$)} 
& KGAT (BERT\textsubscript{BASE})         & -- & -- & 89.0 & 39.6 & 71.2 & 50.3 & -- & -- & -- & -- & 55.2 & -- & 83.6 \\
& CorefBERT\textsubscript{BASE}           & -- & -- & 87.5 & 39.9 & 72.8 & 52.5 & -- & -- & -- & -- & --   & -- & --   \\
& KGAT (RoBERTa\textsubscript{LARGE})     & -- & -- & 91.5 & 60.6 & 74.7 & 60.1 & -- & -- & -- & -- & --   & -- & --   \\
& CorefRoBERTa\textsubscript{LARGE}       & -- & -- & 92.2 & 65.1 & 77.5 & 60.7 & -- & -- & -- & -- & --   & -- & --   \\
\cmidrule(lr){1-2}
\multirow{4}{*}{Reorder ($\epsilon = 5$)} 
& KGAT (BERT\textsubscript{BASE}) & -- & -- & 89.0 & 37.8 & 71.2 & 49.5 & -- & -- & -- & -- & 55.1 & -- & 81.8 \\
& CorefBERT\textsubscript{BASE} & -- & -- & 87.5 & 37.4 & 72.8 & 52.3 & -- & -- & -- & -- & -- & -- & -- \\
& KGAT (RoBERTa\textsubscript{LARGE}) & -- & -- & 91.5 & 47.7 & 74.7 & 54.8 & -- & -- & -- & -- & -- & -- & -- \\
& CorefRoBERTa\textsubscript{LARGE} & -- & -- & 92.2 & 49.9 & 77.5 & 51.4 & -- & -- & -- & -- & -- & -- & -- \\
\cmidrule(lr){1-2}
\multirow{4}{*}{Delete ($\epsilon = 5$)} 
& KGAT (BERT\textsubscript{BASE}) & -- & -- & 89.0 & 38.9 & 71.2 & 49.7 & -- & -- & -- & -- & 60.5 & -- & 79.4 \\
& CorefBERT\textsubscript{BASE} & -- & -- & 87.5 & 39.2 & 72.8 & 52.5 & -- & -- & -- & -- & -- & -- & -- \\
& KGAT (RoBERTa\textsubscript{LARGE}) & -- & -- & 91.5 & 60.8 & 74.7 & 60.7 & -- & -- & -- & -- & -- & -- & -- \\
& CorefRoBERTa\textsubscript{LARGE} & -- & -- & 92.2 & 66.1 & 77.5 & 59.8 & -- & -- & -- & -- & -- & -- & -- \\
\midrule
Imperceptible\textsubscript{Ret} &  &  &  &  &  &  &  &  &  &  &  &  \\ 
\cmidrule(lr){1-2}
Homoglyph ($\epsilon = 5$) 
& KGAT (BERT\textsubscript{BASE}) & -- & -- & 89.0 & 62.3 & 71.2 & 60.5 & -- & -- & -- & -- & 31.5 & -- & 88.9\\
\midrule
\multirow{4}{*}{Lexical Variation}  & KGAT (BERT\textsubscript{BASE}) & -- & -- & 89.0 & 68.9 & 71.2 & 65.4 & -- & -- & -- & -- & 42.1 & -- & 73.6 \\
& CorefBERT\textsubscript{BASE} & -- & -- & 87.5 & 67.7 & 72.8 & 66.6 & -- & -- & -- & -- & -- & -- & -- \\
& KGAT (RoBERTa\textsubscript{LARGE}) & -- & -- & 91.5 & 73.6 & 74.7 & 69.9 & -- & -- & -- & -- & -- & -- & -- \\
& CorefRoBERTa\textsubscript{LARGE} & -- & -- & 92.2 & 74.5 & 77.5 & 71.6 & -- & -- & -- & -- & -- & -- & -- \\
\midrule
\multirow{4}{*}{Contextualized Replace} & KGAT (BERT\textsubscript{BASE}) & -- & -- & 89.0 & 50.7 & 71.2 & 59.7 & -- & -- & -- & -- & 30.3 & -- & 69.3 \\
& CorefBERT\textsubscript{BASE} & -- & -- & 87.5 & 50.0 & 72.8 & 60.8 & -- & -- & -- & -- & -- & -- & --  \\
& KGAT (RoBERTa\textsubscript{LARGE}) & -- & -- & 91.5 & 55.8 & 74.7 & 63.9 & -- & -- & -- & -- & -- & -- & -- \\
& CorefRoBERTa\textsubscript{LARGE} & -- & -- & 92.2 & 55.6 & 77.5 & 64.0 & -- & -- & -- & -- & -- & -- & -- \\
\midrule
\multirow{4}{*}{Omitting Paraphrase}  & KGAT (BERT\textsubscript{BASE}) & -- & -- & 89.0 & 51.0 & 71.2 & 54.3 & -- & -- & -- & -- & 54.4 & -- & 83.8 \\
& CorefBERT\textsubscript{BASE} & -- & -- & 87.5 & 50.7 & 72.8 & 55.8 & -- & -- & -- & -- & -- & -- & -- \\
& KGAT (RoBERTa\textsubscript{LARGE}) & -- & -- & 91.5 & 56.8 & 74.7 & 60.7 & -- & -- & -- & -- & -- & -- & -- \\
& CorefRoBERTa\textsubscript{LARGE} & -- & -- & 92.2 & 55.5 & 77.5 & 58.4 & -- & -- & -- & -- & -- & -- & -- \\
\midrule
\multirow{4}{*}{Omitting Generate}  & KGAT (BERT\textsubscript{BASE}) & -- & -- & 89.0 & 29.9 & 71.2 & 46.8 & -- & -- & -- & -- & 30.9 & -- & 87.9 \\
& CorefBERT\textsubscript{BASE} & -- & -- & 87.5 & 30.2 & 72.8 & 48.9 & -- & -- & -- & -- & -- & -- & --  \\
& KGAT (RoBERTa\textsubscript{LARGE}) & -- & -- & 91.5 & 33.8 & 74.7 & 51.9 & -- & -- & -- & -- & -- & -- & --  \\
& CorefRoBERTa\textsubscript{LARGE} & -- & -- & 92.2 & 31.6 & 77.5 & 47.4 & -- & -- & -- & -- & -- & -- & --  \\
\midrule
\multirow{4}{*}{\begin{tabular}[c]{@{}c@{}}Claim-aligned\\ Re-writing\end{tabular}}  & KGAT (BERT\textsubscript{BASE}) & -- & -- & -- & -- & 71.2 & 38.4\tnote{\textcolor{cyan}{2}} & -- & -- & -- & -- & 94.4\tnote{\textcolor{cyan}{2}} & -- & 1.8\tnote{\textcolor{cyan}{2}} \\
& CorefBERT\textsubscript{BASE} & -- & -- & -- & -- & 72.8 & 36.9 & -- & -- & -- & -- & -- & -- & -- \\
& KGAT (RoBERTa\textsubscript{LARGE}) & -- & -- & -- & -- & 74.7 & 44.9 & -- & -- & -- & -- & -- & -- & -- \\
& CorefRoBERTa\textsubscript{LARGE} & -- & -- & -- & -- & 77.5 & 42.1 & -- & -- & -- & -- & -- & -- & -- \\
\midrule

\multirow{4}{*}{\begin{tabular}[c]{@{}c@{}}Claim-aligned\\ Re-writing\textsubscript{ret} \end{tabular}}  & KGAT (BERT\textsubscript{BASE}) & -- & -- & -- & -- & 71.2 & 43.7\tnote{\textcolor{cyan}{2}} & -- & -- & -- & -- & 99.1\tnote{\textcolor{cyan}{2}} & -- & 1.8\tnote{\textcolor{cyan}{2}} \\
& CorefBERT\textsubscript{BASE} & -- & -- & -- & -- & 72.8 & 43.1 & -- & -- & -- & -- & -- & -- & -- \\
& KGAT (RoBERTa\textsubscript{LARGE}) & -- & -- & -- & -- & 74.7 & 48.8 & -- & -- & -- & -- & -- & -- & -- \\
& CorefRoBERTa\textsubscript{LARGE} & -- & -- & -- & -- & 77.5 & 47.6 & -- & -- & -- & -- & -- & -- & -- \\
\midrule

\multirow{4}{*}{\begin{tabular}[c]{@{}c@{}}Supporting Generation\end{tabular}} & KGAT (BERT\textsubscript{BASE}) & -- & -- & -- & -- & 71.2 & 42.0\tnote{\textcolor{cyan}{2}} & 72.4 & 32.2\tnote{\textcolor{cyan}{2}} & -- & -- & 85.7\tnote{\textcolor{cyan}{2}} & -- & 3.8\tnote{\textcolor{cyan}{2}} 
\\
& CorefBERT\textsubscript{BASE} & -- & -- & -- & -- & 72.8 & 39.7 & 72.8 & 34.5 & -- & -- & -- & -- & -- \\
& KGAT (RoBERTa\textsubscript{LARGE}) & -- & -- & -- & -- & 74.7 & 43.2 & 68.8 & 25.6 & -- & -- & -- & -- & -- \\
& CorefRoBERTa\textsubscript{LARGE} & -- & -- & -- & -- & 77.5 & 45.2 & 70.0 & 30.0 & -- & -- & -- & -- & -- \\
\midrule

\multirow{4}{*}{\begin{tabular}[c]{@{}c@{}}Omission Generation\end{tabular}} & BERT & -- & -- & -- & -- & -- & -- & -- & -- & 87.16\tnote{\textcolor{olive}{3}} & 59.51\tnote{\textcolor{olive}{3}} & -- & -- & -- \\
& RoBERTa & -- & -- & -- & -- & -- & -- & -- & -- & 88.69\tnote{\textcolor{olive}{3}} & 59.10\tnote{\textcolor{olive}{3}} & -- & -- & -- \\
& ALBERT  & -- & -- & -- & -- & -- & -- & -- & -- & 86.67\tnote{\textcolor{olive}{3}} & 63.00\tnote{\textcolor{olive}{3}} & -- & -- & -- \\
& Ensemble\tnote{\textcolor{red}{1}} & -- & -- & -- & -- & -- & -- & -- & -- & 88.81\tnote{\textcolor{olive}{3}} & 61.36\tnote{\textcolor{olive}{3}} & -- & -- & -- \\
\midrule
\multirow{2}{*}{\begin{tabular}[c]{@{}c@{}}Neutral Noise \end{tabular}} & BM25 retriever & -- & -- & -- & -- & -- & -- & -- & -- & -- & -- & -- & 35.17\tnote{\textcolor{green}{4}} & -- \\
& QR-BERT retriever & -- & -- & -- & -- & -- & -- & -- & -- & -- & -- & -- & 54.10\tnote{\textcolor{green}{4}} & -- \\
\bottomrule
\end{tabular}%
}                
\begin{tablenotes}

\tiny
\item \textbf{Acc}: Accuracy; \textbf{Rec$_{\text{Evd}}^{\text{Adv}}$}: Recall of adversarial evidence; \textbf{Rec$_{\text{Evd}}^{\text{Gold}}$}: Recall of gold evidence after attack; \textbf{→ NEI}: the ratio of verdict predictions that shift to NEI; \\ \textbf{Org}: Original dataset (before attack); \textbf{Mod}: Modified dataset (after attack); \textbf{--}: This metric was not assessed.
\item \underline{Test FC systems}: KGAT~\cite{liu-etal-2020-fine}; BERT~\cite{devlin-etal-2019-bert}, CorefBERT, CoreRoBERTa~\cite{ye-etal-2020-coreferential}; MLA~\cite{kruengkrai-etal-2021-multi}; RoBERTAa~\cite{liu2019roberta};\\ ALBERT~\cite{lan2020ALBERT}; BM25~\cite{robertson1994okapi}; QR-BERT~\cite{samarinas2020latent}.
\item[\textcolor{red}{1}] An ensemble of three FC models, i.e., BERT, RoBERTa, and ALBERT, in \cite{atanasova2022fact}.
\item[\textcolor{cyan}{2}] Statistics for Acc are reported under stance filtering for both Claim-aligned Re-writing and Supporting Generation attacks, and under retrieval filtering for the Claim-aligned\\ Re-writing$_{\text{Ret}}$ attack.
\item[\textcolor{olive}{3}] Statistics for Macro-F1 score on claim verification results before and after attack.
\item[\textcolor{green}{4}] Statistics for Recall@5 on gold evidence retrieval with adversarial evidence included.

\end{tablenotes}
\end{threeparttable}
\caption{Evaluation of adversarial evidence attacks’ performance against test FC systems on FEVER 1.0.}
\label{tab:adv_evidence_eval}
\end{table*}

\end{document}